\documentclass[10pt,twocolumn,letterpaper]{article}

\usepackage{iccv}              

\definecolor{iccvblue}{rgb}{0.21,0.49,0.74}
\definecolor{GammaColor}{rgb}{0.5,0,0.7}

\usepackage[pagebackref,breaklinks,colorlinks,allcolors=iccvblue]{hyperref}
\usepackage{multirow}

\def\paperID{7593} 
\def\confName{ICCV}
\def\confYear{2025}

\title{Learning an Implicit Physics Model for Image-based Fluid Simulation}

\author{Emily Yue-Ting Jia\quad
Jiageng Mao\quad
Zhiyuan Gao\quad
Yajie Zhao\quad
Yue Wang\\
University of Southern California\\
{\tt\small \{eyjia, jiagengm, gaozhiyu, yue.w\}@usc.edu} \quad
{\tt\small zhao@ict.usc.edu}
}

\begin{document}

\twocolumn[{%
\renewcommand\twocolumn[1][]{#1}%
\maketitle
\begin{center}
    \vspace{-15pt}
    \centering
    \captionsetup{type=figure}
    \includegraphics[width=\textwidth, trim=0 0 0 0, clip]{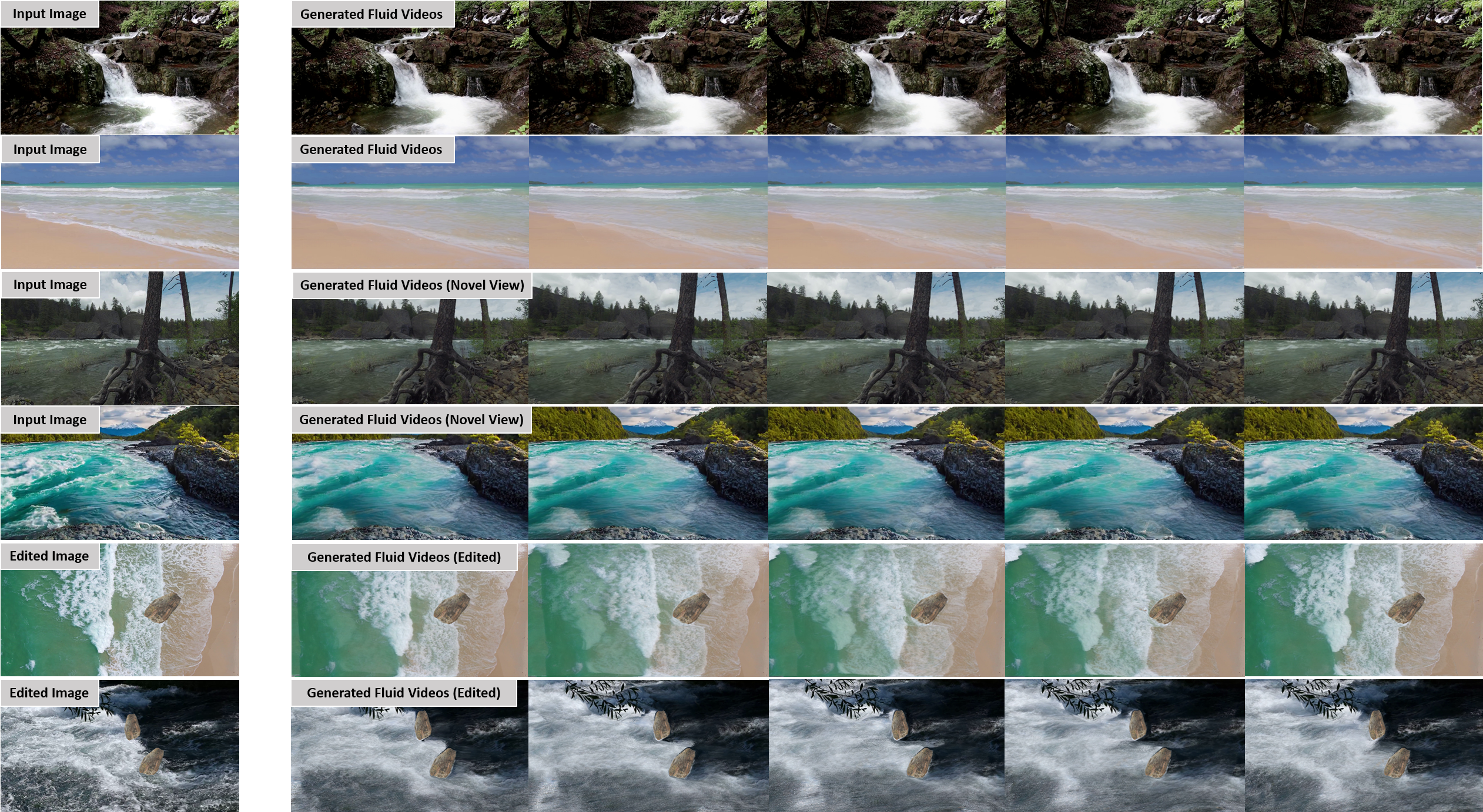}
    \vspace{-15pt}
    \captionof{figure}{
    Our goal is to create realistic fluid animations that are consistent with basic physics laws from a single still image. Our methods handle the obstructions added in the input images and generate realistic fluid animation videos from novel views and moving cameras. }
\label{fig:teaser}
\end{center}
}]

\begin{abstract}
Humans possess an exceptional ability to imagine 4D scenes, encompassing both motion and 3D geometry, from a single still image. This ability is rooted in our accumulated observations of similar scenes and an intuitive understanding of physics. In this paper, we aim to replicate this capacity in neural networks, specifically focusing on natural fluid imagery. Existing methods for this task typically employ simplistic 2D motion estimators to animate the image, leading to motion predictions that often defy physical principles, resulting in unrealistic animations. Our approach introduces a novel method for generating 4D scenes with physics-consistent animation from a single image. We propose the use of a physics-informed neural network that predicts motion for each surface point, guided by a loss term derived from fundamental physical principles, including the Navier-Stokes equations. To capture appearance, we predict feature-based 3D Gaussians from the input image and its estimated depth, which are then animated using the predicted motions and rendered from any desired camera perspective. Experimental results highlight the effectiveness of our method in producing physically plausible animations, showcasing significant performance improvements over existing methods. Our project page is \href{https://physfluid.github.io/}{https://physfluid.github.io/}.

\end{abstract}    
\section{Introduction}
\label{sec:intro}
Given a single image of natural fluids, humans can vividly imagine the surrounding scene and the potential movements within it. For instance, an image of a stream might evoke a tranquil walk along its crystal-clear, shallow waters in a peaceful forest, accompanied by the gentle sound of water flowing over smooth stones. Similarly, a photo of a waterfall cascading down a rugged cliff can create a sensory experience, bringing to mind the powerful roar of water crashing into a pool below and the awe of witnessing nature’s grandeur. This imaginative ability to construct a 4D scene, blending spatial and temporal elements from a single image, is essential to the human experience. 

This paper focuses on replicating the human imaginative ability to construct 4D natural fluid scenes using neural networks. We identify this ability as a video generation problem, where the goal is to enable the model to generate videos that depict both fluid motion and camera movement, given a single image and a specified camera trajectory.

Although such an ability is commonly possessed by humans, replicating this imaginative capacity in computational models is a significant challenge, even when we focus only on fluids. One possible approach is leveraging simulation. However, traditional fluid simulation demands precise knowledge of the boundary geometry and physical properties such as mass, initial velocity, and external forces. Such detailed information, however, is difficult to infer solely from a single image. Another potential approach is to learn fluid motion solely from data priors. However, due to the complexity of fluid dynamics, these methods~\cite{cai2022diffdreamer, 3d-cinemagraphy, holynski} often produce animations that appear unrealistic. For example, the fluids in the animation may penetrate the occlusions, as these methods have no understanding of collisions. 

In this paper, we propose physics-informed neural dynamics, a data-driven approach that predicts physically grounded flow dynamics from a single fluid image. The core of this method is a conditional physics-informed neural network that learns from both data priors and additional guidance from fundamental physics laws. Our approach addresses the challenge of predicting detailed physical properties by directly estimating the 3D velocity field from the input image. By leveraging both data priors and physical guidance, our physics-informed neural dynamics produces fluid animations that are both visually compelling and physically accurate. Experiments show that our model effectively captures the interaction between fluids and boundaries while successfully adapting to changes in boundary geometry.

While our physics-informed neural dynamics predicts physically accurate fluid motions, visualizing these motions in videos requires transforming the input image into a representation suitable for animation. We adopt 3D Gaussians~\cite{3dgs} as the representation, as they facilitate novel-view synthesis and are also easy to animate. Using the velocity field predicted by the physics-informed neural dynamics, the 3D Gaussian representation—obtained by lifting pixels with their relative depth—can be animated by displacing each Gaussian kernel according to its corresponding velocity. The displaced kernels can be rendered from any camera pose to produce the desired video frames. By combining our physics-informed neural dynamics with the 3D Gaussian representation, our method generates physically plausible videos of natural fluid motions from any specified camera trajectory, effectively replicating the human ability to envision and construct a 4D scene from a single image. Further comparison with baseline methods demonstrates that our method achieves higher visual quality from both statistical evaluations and human judgment. Furthermore, experiments on boundary editing tasks show that our method successfully adapts to boundary changes, whereas the baseline method generates unrealistic animations.

To summarize, our contributions are:

$\bullet$ We introduce a novel method to generate a 4D scene from a single image, producing physically plausible videos that capture both fluid motion and camera movement.

$\bullet$ We propose physics-informed neural dynamics, a data-driven approach that can predict physics-grounded flow dynamics from a single image of natural fluids.

$\bullet$ Experiments show that our framework surpasses baseline methods, producing more realistic animations, particularly in image editing tasks where the boundary is changed. 

\begin{figure}
    \centering
    \includegraphics[width=\linewidth]{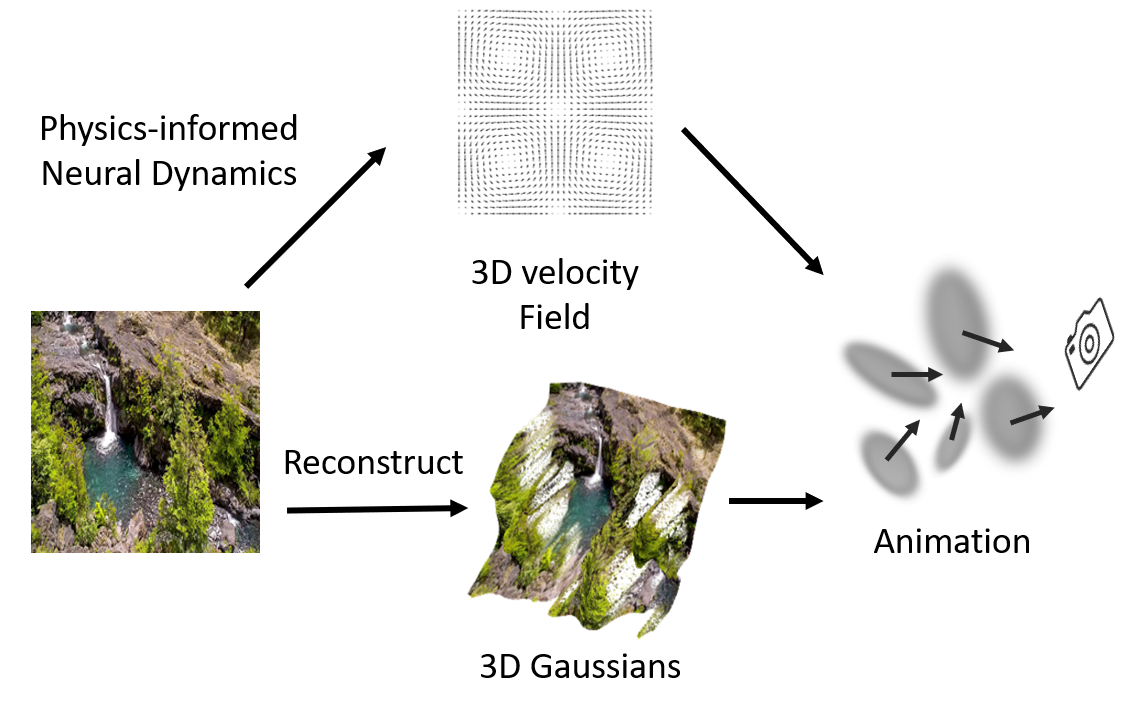}
    \vspace{-8pt}
    \caption{An overview of our proposed method. Our method consists of two parts: the {\bf Physics-informed Neural Dynamics}, which predicts the fluid motions, and the {\bf Animation Module}, which reconstructs the 3D geometry of the scene.}
    \label{fig:overview}
    \vspace{-10pt}
\end{figure}

\section{Related Works}
\label{sec:RW}

\paragraph{Single Image Animation:} Many studies explore video generation from a single image. Diffusion-based methods~\cite{ho2022imagen,SVD,ho2022video,singer2022make,hong2022cogvideo,yang2024cogvideox} generate videos from images or text prompts. However, these methods require careful prompting to effectively control, and the animations produced by these methods often lack physical plausibility and 3D consistency. Recent works~\cite{cai2022diffdreamer,gao2024cat3d,yu2024viewcrafter,liang2024wonderland,xie2024sv4d,yu2024wonderjourney} add 3D constraints for multi-view consistency but mostly focus on static scenes. Others~\cite{liu2024physgen,tan2024physmotion,zhang2024physdreamer,xu2024motion} incorporate physics for realistic animations, yet none address fluid animation.


Besides diffusion-based approaches, alternative methods have also been explored.~\cite{chuang2005animating,jhou2015animating} suggest using motions to guide animation generation. Following them, motion transfer methods~\cite{chan2019everybody,dosovitskiy2015flownet,liu2019liquid,ren2020deep,siarohin2019first} require reference videos and are unsuitable for our task. Other methods~\cite{endo2019animating,3d-cinemagraphy,mahapatra2022controllable, holynski} employ learnable motion estimators.~\cite{endo2019animating, mahapatra2022controllable, holynski} adopt 2D feature warping with the predicted motion maps, while~\cite{3d-cinemagraphy} transforms the image to 3D point clouds and animates them by lifting the 2D motion maps to 3D.~\cite{make-it-4d} further extends ~\cite{3d-cinemagraphy} to support longer camera trajectories by using Diffusion model to inpaint the invisible areas. However, lacking the knowledge of physics, these methods produce unrealistic animations. In contrast, our approach integrates both physics and data priors for more realistic results.

\noindent\textbf{Physics-based Fluid Dynamics:}
Physics simulation offers another approach to animation, particularly for dynamic elements in images. Traditional simulation relies on explicit modeling, requiring precise initial states and 3D geometry~\cite{grzeszczuk1998neuroanimator,peng2018deepmimic,SLR-SFS}, limiting its scalability. With the development of physics-informed neural networks~\cite{pinn}, many works try to incorporate them in fluid simulation. Most efforts~\cite{deng2023fluid,wandel2020learning,wandel2021teaching,tompson2017accelerating,geneva2020modeling,li2022graph} focus on reducing the computational costs of simulation by using neural networks as PDE solvers. Recently,~\cite{chu2022smoke,yu2023hyfluid,guan2022neurofluid,franz2023learning,franz2021global,zang2020tomofluid} use implicit neural representations to recover fluid dynamics from videos.~\cite{guan2022neurofluid} requires ground truth reconstructed initial fluid states, while ~\cite{chu2022smoke,yu2023hyfluid,franz2023learning,franz2021global} use physics-informed neural network to recover the initial states. However, these methods are optimized per-scene.  In contrast, our method extends these methods and learns generalizable fluid dynamics from large-scale videos.


\section{Methods}
\label{sec:Methods}
\begin{figure*}[t!]
  \centering  
  \includegraphics[width=\linewidth]{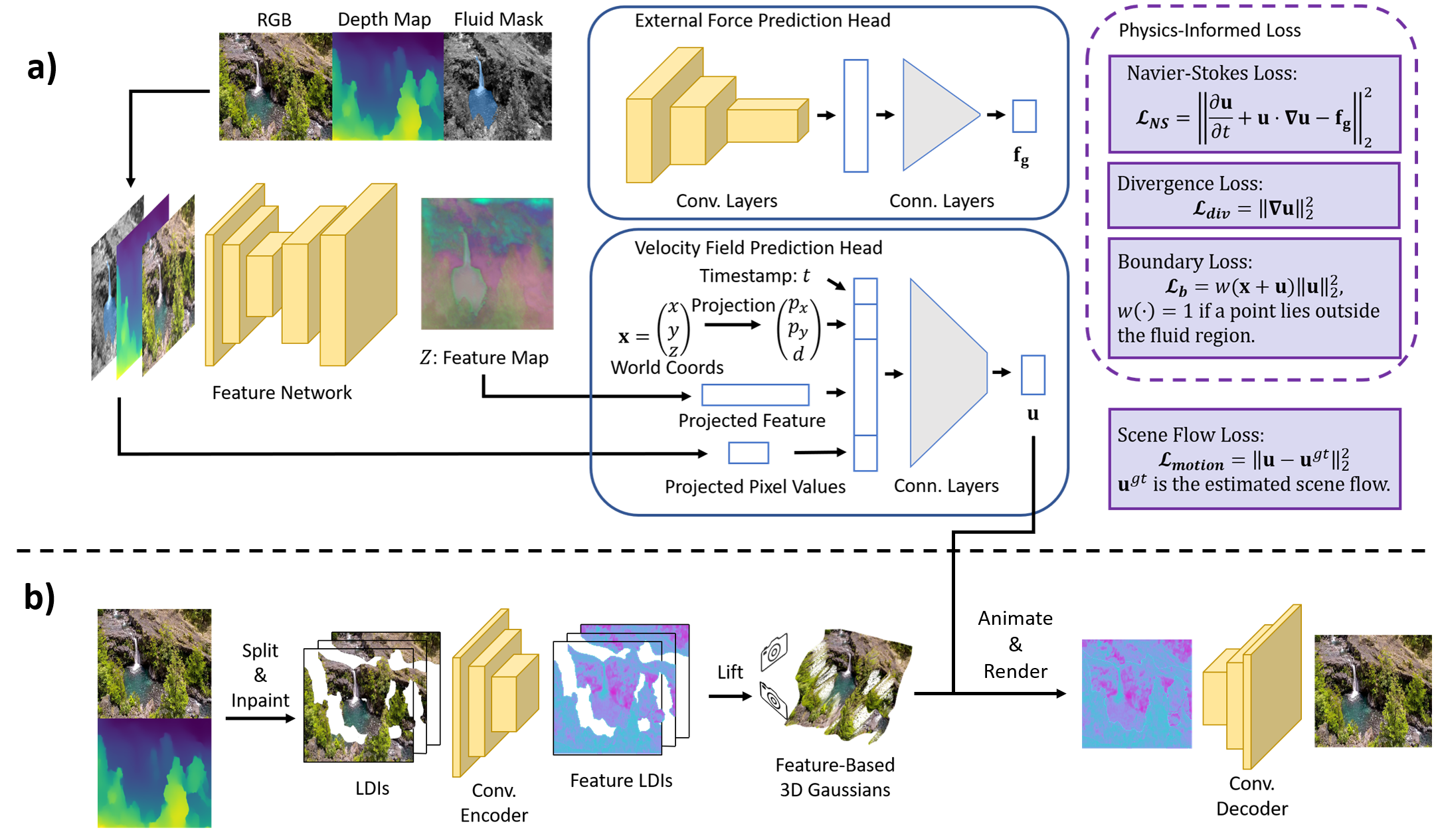}
  \vspace{-15pt}
  \caption{A detailed overview of the {\bf Physics-Informed Neural Dynamics} and the {\bf Animation Module}. The {\bf Physics-Informed Neural Dynamics} (top) takes an RGB image, depth map, and additional fluid area mask as input to predict 3D velocity field $(\mathbf{x}, t) \rightarrow \mathbf{u}$ and the external force $\mathbf{f}_g \in \mathbb{R}^3$. The velocity prediction is supervised by both estimated 3D scene flows and an additional physics-informed loss to ensure realistic motions. The predicted external force $\mathbf{f}_g$ is only used to compute the physics-informed loss. The {\bf Animation Module} (bottom) converts an image into Layered Depth Images (LDIs) and then 3D Gaussians, which can be animated and rendered in novel views. }
  \label{fig:pipeline}
  \vspace{-10pt}
\end{figure*}

\subsection{Overview}
\label{subsec:overview}
Given a single natural fluid image $I_0$ and a camera trajectory $\{P_t\}_{t=1}^T$, our goal is to generate a $T$-frame video $\{\hat{I}_t\}_{t=1}^T$ that captures both fluid motion and camera movement. Our method consists of two components: physics-informed neural dynamics, which learns a 3D fluid motion prior with the guidance of physics laws, and the animation module, which applies the predicted motions from the physics-informed neural dynamics to animate the 2D image. A detailed overview of our pipeline is shown in \Cref{fig:pipeline}. 

We adopt 3D Gaussians~\cite{3dgs} as the fluid particle representation. 
The physics-informed neural dynamics predicts 3D velocity fields $\mathbf{u}: (\mathbf{x}, t) \rightarrow  \mathbb{R}^3$ from the input image $I_0$, where $\mathbf{x}$ represents the centers of 3D Gaussians, and $t$ denotes the timestamp. In the animation module, we animate the 3D Gaussians based on the learned velocities from $\mathbf{u}$. Finally, the animated Gaussians are rendered to produce the predicted frame  $\hat{I}_t$ in the video.


\subsection{Fluid Representation}
\label{subsec:gs}
Our goal is to generate videos from a single image that capture both fluid motion and camera movement. To accomplish this, we need a fluid representation that is well-suited for animation and aligns with our physics-informed neural dynamics. This representation must be in 3D, as generating videos with camera movements requires the knowledge of 3D geometry. To this end, we adopt pixel-aligned 3D Gaussians~\cite{3dgs} as our fluid representation. 3D Gaussians can be easily derived from a single image by lifting pixels into 3D space using depth information. Additionally, it facilitates animation by simply displacing the centers of each Gaussian. Unlike 3D point clouds~\cite{3d-cinemagraphy}, 3D Gaussians offer natural blending in empty regions, effectively addressing the issues of holes that arise during point cloud rasterization.

\subsubsection{3D Gaussian Splatting}
 3D Gaussians, proposed by~\cite{3dgs}, represent a scene with a set of anisotropic 3D Gaussians. Each 3D gaussian $G(\mathbf{x})$ is defined by its center $\mu$ and a 3D covariance matrix $\mathbf{\Sigma}$: 
\begin{equation}
    G(\mathbf{x}) = \exp{(-\frac{1}{2}(\mathbf{x}-\mu)^T \mathbf{\Sigma}^{-1} (\mathbf{x}-\mu)}).
\end{equation}
The covariance matrix $\mathbf{\Sigma}$ can be further decomposed into $\mathbf{\Sigma} = \mathbf{R}\mathbf{S}\mathbf{S}^T\mathbf{R}^T$, where $\mathbf{S}$ is the diagonal scale matrix and $\mathbf{R}$ is the rotation matrix. Each Gaussian kernel is also associated with a learnable opacity $o$ and a learnable color function $c(d)$. $d$ is the viewing direction. $c(d)$ is represented by spherical harmonics. Given any camera pose, 3D Gaussians can be rendered using a tile-based rasterizer:
\begin{equation}
    \hat{I}(v) = \sum_{i\in N} c_i \alpha_i \prod_{j=1}^{i-1} (1-\alpha_j),
\label{eq:gs}
\end{equation}
where $\hat{I}(v)$ is the color of pixel $v$, $N$ is the total number of visible Gaussian kernels in the tile, and $c_i$ is the color of the i-th Gaussian viewing from the given camera pose. $\alpha_i$ is computed using $\alpha_i = o_iG^p(\mathbf{x})$, where $G^p(\mathbf{x})$ is the function acquired by projecting $G(\mathbf{x})$ to the 2D image plane.

\subsubsection{3D Gaussians from Single Image}
We adopt a similar approach to~\cite{3d-cinemagraphy} to derive feature-based 3D Gaussians $G_0: \{(\mathbf{x}^i_0, \mathbf{R}^i_0, \mathbf{S}^i_0, \alpha^i_0, z^i_0)\}^M$ from the input image $I_0$, where $z^i$ denotes the feature tensor for the i-th gaussian kernel. Specifically, we first convert the input image into Layered Depth Images (LDIs) by applying agglomerative clustering on the monocular depth map. Each LDI layer is then inpainted and processed to extract features. The processed LDIs are subsequently transformed into feature-based 3D Gaussians by lifting each LDI with its monocular depth. The opacity parameters $\alpha^i$ are initialized based on the confidence predicted by the inpainting network, and the scale parameters $\mathbf{S}^i$ are set proportional to their distance from the image plane. 

\subsection{Physics-Informed Neural Dynamics}
\label{subsec:physics}
Our physics-informed neural dynamics learns a 3D space prior for fluid motions from real-world videos, augmented by the guidance of physical laws. Our key insight is that by leveraging physics-informed neural dynamics, we can generate fluid animations that are both visually compelling and physics-aware. The core of this part is to train a conditional physics-informed neural network to predict the velocity field $\mathbf{u}: (\mathbf{x}, t)\rightarrow \mathbb{R}^3$ for fluids based on the input image $I_0$. This network is supervised with dense scene flows and extra loss terms derived from physics laws. 

\subsubsection{Injecting Physical Laws into Neural Dynamics}
\label{subsubsec:physic}
In this section, we introduce the governing physical laws for fluid simulation and then elaborate on how to leverage the physical laws to guide the learning of neural dynamics.

In our setting, we assume all the fluids to be incompressible, whose motion follows the incompressible Navier-Stokes Equations:
\begin{equation}
    \frac{\partial \mathbf{u}}{\partial t} + \mathbf{u} \cdot \mathbf{\nabla} \mathbf{u} = -\frac{1}{\rho} \mathbf{\nabla} p + \nu \mathbf{\nabla} \cdot \mathbf{\nabla} \mathbf{u} + \frac{1}{\rho}\mathbf{f}_g,
\label{eq:ns_eq}
\end{equation}
\begin{equation}
    \mathbf{\nabla} \cdot \mathbf{u} = 0,
\label{eq:div}
\end{equation}
where $\mathbf{u}$ is the velocity field, $p$ is the pressure field $p: (\mathbf{x}, t) \rightarrow \mathbb{R}$, $\mathbf{f}_g\in \mathbb{R}^3$ is the external forces, $\rho$ is the fluid density and $\nu$ is the kinematic viscosity. Eq. \ref{eq:div} is also known as the incompression constraint. 

When deriving the loss terms for our network, we apply two simplifications in \Cref{eq:ns_eq}. Firstly, following\cite{chu2022smoke, yu2023hyfluid}, we assume inviscid fluids  and omit the viscosity term, $\nu \mathbf{\nabla} \cdot \mathbf{\nabla} \mathbf{u}$. Secondly, for the pressure term,~\cite{chu2022smoke} noted that adding additional networks for pressure fields would significantly increase the degrees of freedom in optimization, hindering the learning when ground truth pressures are not available. Therefore, we also exclude the pressure term.

As stated by~\cite{chu2022smoke}, the simplifications are``with valid assumptions'': the networks will find possible solutions with minimal influence from pressure difference and viscosity. 

The simplified Navier-Stokes Equations go as:
\begin{equation}
    \frac{\partial \mathbf{u}}{\partial t} + \mathbf{u} \cdot \mathbf{\nabla} \mathbf{u} = \mathbf{f}_g, 
\end{equation}
We also eliminate the $\frac{1}{\rho}$ before $\mathbf{f}_g$ as it could be learned together with $\mathbf{f}_g$ using neural networks. 

To acquire physics-grounded animation, we want the velocity field $\mathbf{u}$, generated by our physics-informed neural dynamics, to follow the simplified Navier-Stokes equations. This is achieved by minimizing the $L_2$-norm between the two sides of the equation:
\begin{equation}
    \mathcal{L}_{NS} = \lVert\frac{\partial \mathbf{u}}{\partial t} + \mathbf{u} \cdot \mathbf{\nabla} \mathbf{u} - \mathbf{f}_g \rVert_2^2,
\end{equation}
\begin{equation}
    \mathcal{L}_{div} = \lVert \mathbf{\nabla} \cdot \mathbf{u} \rVert_2^2.
\end{equation}

During training, we compute $\mathcal{L}_{NS}$ and $\mathcal{L}_{div}$ over all the fluid surface points to provide guidance for our network. 

Fluids in the image also collide with the boundaries, \eg rocks and river banks. We assume all boundaries are impermeable and enforce a Dirichlet (no-through) boundary condition on the velocity field.:
\begin{equation}
    \mathbf{u}(\mathbf{x}, t) \cdot \mathbf{n}_{\Gamma}(\mathbf{x}) = 0, \mathbf{x}\in \Gamma,
\end{equation}
where $\Gamma$ denotes the boundary and $\mathbf{n}_{\Gamma}(\mathbf{x})$ denotes the surface normal at boundary point $\mathbf{x}$. 

To enforce this constraint during training, we sample additional points near the boundaries and calculate their velocities. Since estimating accurate boundary normals from a single image is challenging, we omit $\mathbf{n}_{\Gamma}$ and penalize the $L_2$-norm of velocities for points incorrectly predicted to exit the fluid region. Formally, for all the points that stay within the fluid region in the ground truth, we apply the following penalty to enforce the boundary condition:
\begin{equation}
        \mathcal{L}_{b} = w(\mathbf{x} + \mathbf{u}) \lVert \mathbf{u}\rVert_2^2,  \\
\end{equation}
where $w(\cdot) \rightarrow \{0, 1\}$ is a binary indicator, with $w(\cdot) = 1$ indicating that a point lies outside the fluid region. In practice, the ground truth movement is derived from the optical flow maps, details of which are explained in Sec. \ref{subsec:optimization}.



The overall physics-aware loss is:
\begin{equation}
    \mathcal{L}_{physics} = \lambda_{NS}  \mathcal{L}_{NS} + \lambda_{div} \mathcal{L}_{div} + \mathcal{L}_{b}.
\end{equation}

\subsubsection{Network Architecture}
\label{subsubsec:network_physic}
The core of our physics-informed neural dynamics is a conditional physics-informed neural network that predicts the velocity field $\mathbf{u}$ based on the input image $I_0$. Our network begins with a convolutional network to extract the feature map $Z$ from the input image $I_0$. These extracted features then serve as the conditions for an MLP $\mathbf{u}_{\Theta}: (\mathbf{x}, t; Z(\mathbf{x})) \rightarrow \mathbb{R}^3$ to predict the velocity for each query. As mentioned in Sec. \ref{subsubsec:physic}, our network also includes an extra head that takes the feature map $Z$ and outputs the external force $\mathbf{f}_g\in \mathbb{R}^3$ for the input image. This extra head is only used during training, where its output $\mathbf{f}_g$ is used to compute the physics-aware loss. 

\noindent\textbf{Flow Hints}: Similar to~\cite{3d-cinemagraphy,holynski,SLR-SFS}, our method can be extended to accept sparse flow hints. An example of the hints can be found in our supplementary materials.

\subsection{Animation}
In Sec. \ref{subsec:gs} and \ref{subsec:physics}, we explain how to acquire the 3D Gaussians $G_0: \{(\mathbf{x}^i_0, \mathbf{R}^i_0, \mathbf{S}^i_0, \alpha^i, z^i_0)\}^M$ and the velocity field $\mathbf{u}_\Theta(\mathbf{x}, t;Z(\mathbf{x}))$ from the input image $I_0$.
With $G_0$ and $\mathbf{u}_\Theta$, we can now animate $G_0$ to get the desired frame $\hat{I}_t$. This is done by iteratively displacing the Gaussian kernels with the predicted velocities. Specifically, given the 3D Gaussians $G_t$ at time $t$, the 3D Gaussian representation at $t+1$ is obtained by displacing the kernel center $\mathbf{x}^i_t$ by:
\begin{equation}
    \mathbf{x}^i_{t+1} = \mathbf{x}^i_t+ \mathbf{u}_\Theta(\mathbf{x}^i_t, t;Z(\mathbf{x}^i_t)).
\label{eq:animate}
\end{equation}

The displaced 3D Gaussians are rendered into 2D feature maps by replacing $c_i$ in Eq. \ref{eq:gs} with the feature vectors $z^i_t$ of the Gaussian kernel. These feature maps are then processed by a decoder to generate the predicted frame $\hat{I}_t$. 

When the fluid points move forward, there will be holes forming in the empty area, leading to visual artifacts. To address this issue, we follow the approach of~\cite{3d-cinemagraphy,holynski} and employ the symmetric splatting method. 



\begin{table*}
    \centering
    \small
    \begin{tabular}{|c|ccc|ccc|ccc|c|}
        \hline
            \multicolumn{1}{|c|}{} & \multicolumn{3}{c|}{All Regions} & \multicolumn{3}{c|}{Fluid Regions} & \multicolumn{3}{c|}{Other Regions} & \multicolumn{1}{|c|}{} \\
            Method & PSNR↑ & SSIM↑ & LPIPS↓ & PSNR↑ & SSIM↑ & LPIPS↓ & PSNR↑ & SSIM↑ & LPIPS↓ & VMAF↑\\
        \hline
        Holynski \etal~\cite{holynski} & 22.11 & 0.64 & 0.34 & 26.10 & 0.82 & 0.18 & 24.50 & 0.72 & 0.32 & 38.88 \\
        3D-Cinemagraphy~\cite{3d-cinemagraphy} & 22.81 & 0.70 & 0.22 & 26.30 & 0.82 & 0.15 & 24.25 & 0.75 & 0.27 & 40.74 \\
        Ours & {\bf 24.98} & {\bf 0.78} & {\bf 0.20} & {\bf 27.50} & {\bf 0.86} & {\bf 0.13} & {\bf 25.92} & {\bf 0.81} & {\bf 0.12} & {\bf 44.40} \\
        \bottomrule
    \end{tabular}
    \vspace{-6pt}
    \caption{Quantitative results of generating videos from the input view on Holynski \etal~\cite{holynski} validation set. We also report the performance on different regions of the input image separately to demonstrate that our method excels in both fluid animation generation and background reconstruction. We replace the uninterested regions with ground truth when computing the metrics for different regions. }
    \vspace{-10pt}
    \label{tab:orig}
\end{table*}

\begin{table*}
  \centering
  \small
  \begin{tabular}{|c|ccc|ccc|ccc|c|}
    \hline
    \multicolumn{1}{|c|}{} & \multicolumn{3}{c|}{All Regions} & \multicolumn{3}{c|}{Fluid Regions} & \multicolumn{3}{c|}{Other Regions} & \multicolumn{1}{|c|}{} \\
    Method & PSNR↑ & SSIM↑ & LPIPS↓ & PSNR↑ & SSIM↑ & LPIPS↓ & PSNR↑ & SSIM↑ & LPIPS↓ & VMAF↑\\
    \hline
    2D Anim→NVS* & 21.12 & 0.63 & 0.29 & - & - & - & - & - & - & - \\
    NVS→2D Anim* & 21.97 & 0.70 & 0.28 & - & - & - & - & - & - & - \\
    3D-Cinemagraphy~\cite{3d-cinemagraphy} & 22.46 & 0.67 & 0.24 & 26.91 & 0.83 & \textbf{0.16} & 25.00 & 0.79 & 0.19 & 37.20 \\
    Make-it-4D~\cite{make-it-4d} & 21.40 & 0.55 & 0.31 & 26.72 & 0.83 & 0.17 & 23.50 & 0.66 & 0.26 & 22.36\\
    Ours & {\bf 24.34} & {\bf 0.76} & {\bf 0.21} & \textbf{27.62} & \textbf{0.86} & 0.17 & \textbf{25.46} & \textbf{0.80} & \textbf{0.18} & \textbf{45.71} \\
    \bottomrule
  \end{tabular}

  \vspace{-8pt}
  \caption{Quantitative results of generating videos from novel views on Holynski \etal~\cite{holynski} validation set. The results in lines with '*' are retrieved from 3D-Cinemagraphy~\cite{3d-cinemagraphy}.}
  \vspace{-15pt}
  \label{tab:sep_novel}
\end{table*}

\subsection{Learning}
\label{subsec:optimization}
The physics-informed neural dynamics and the animation part are trained separately in our method. The reason for separate training is to avoid inaccurate velocity predictions from hindering the optimization of the animation part. 

\subsubsection{Learning Physic-Informed Neural Dynamics}
Physics-informed neural networks are typically trained using only physics-based loss functions. However, in our case, we aim to learn a data-driven prior to recover velocity fields from images, rather than fitting a single fluid scene, which makes optimization more challenging. To address this, we also supervise our network with scene flows extracted from videos. The scene flows will provide strong supervision for image-specific motions, while the physics-aware loss ensures adherence to physical laws. Specifically, we first compute 2D optical flow maps for consecutive frame pairs using~\cite{raft}. Then, utilizing the monocular depth maps, the dense flow maps are lifted to 3D scene flows, which serve as supervision for our network. Formally, for all points in the 3D scene flow, we minimize the End Point Error(EPE) between the predicted velocity and the scene flow vector: 
\begin{equation}
    \mathcal{L}_{motion} = \lVert \mathbf{u} - \mathbf{u}^{gt} \rVert_2.
\end{equation}

The overall training loss for this part is: 
\begin{equation}
    \mathcal{L} = \mathcal{L}_{motion} + \lambda_{physics} \mathcal{L}_{physics}.
\end{equation}

\subsubsection{Learning Photo-realistic Rendering}
To train the networks in the animation part, we supervised the predicted intermediate frame $\hat{I}_t$ with both L1 loss and VGG loss~\cite{lpips,vgg} to achieve better image quality: 
\begin{equation}
    \mathcal{L}_{animation} = \sum_{t}L_1(\hat{I}_t, I_t) + \lambda_{vgg}\mathcal{L}_{vgg}(\hat{I}_t, I_t).
\end{equation}

To enable the ability to view the animation in novel views, similar to~\cite{3d-cinemagraphy}, we use~\cite{3d-photo} to generate novel view images for the first frame $I_0$ of every clip. The generated images serve as pseudo ground truth to supervise our animation part. During training, we render the predicted 3D Gaussians in novel view and supervise them using L1 loss:
\begin{equation}
   \mathcal{L}_{novel\_view} = L_1(\hat{I}_{novel}, I_{novel}).
\end{equation}

The overall supervision of our animation branch is: 
\begin{equation}
     \mathcal{L} = \mathcal{L}_{animation} + \lambda_{novel\_view}L_{novel\_view}.
\end{equation}

\begin{figure*}[t]
  \centering
  \begin{subfigure}{0.15\linewidth}
    \includegraphics[height=.66\linewidth]{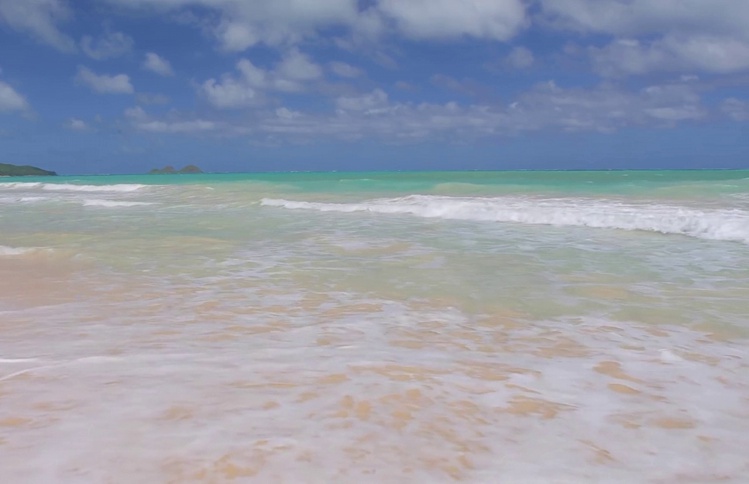}
    \includegraphics[height=.66\linewidth]{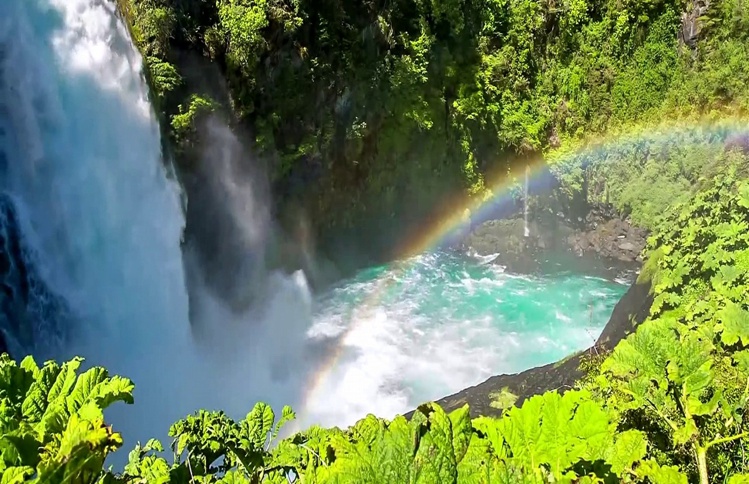}
    \caption{Ground Truth}
    \label{fig:orig-gt}
  \end{subfigure}
  \begin{subfigure}{0.25\linewidth}
    \includegraphics[width=\linewidth]{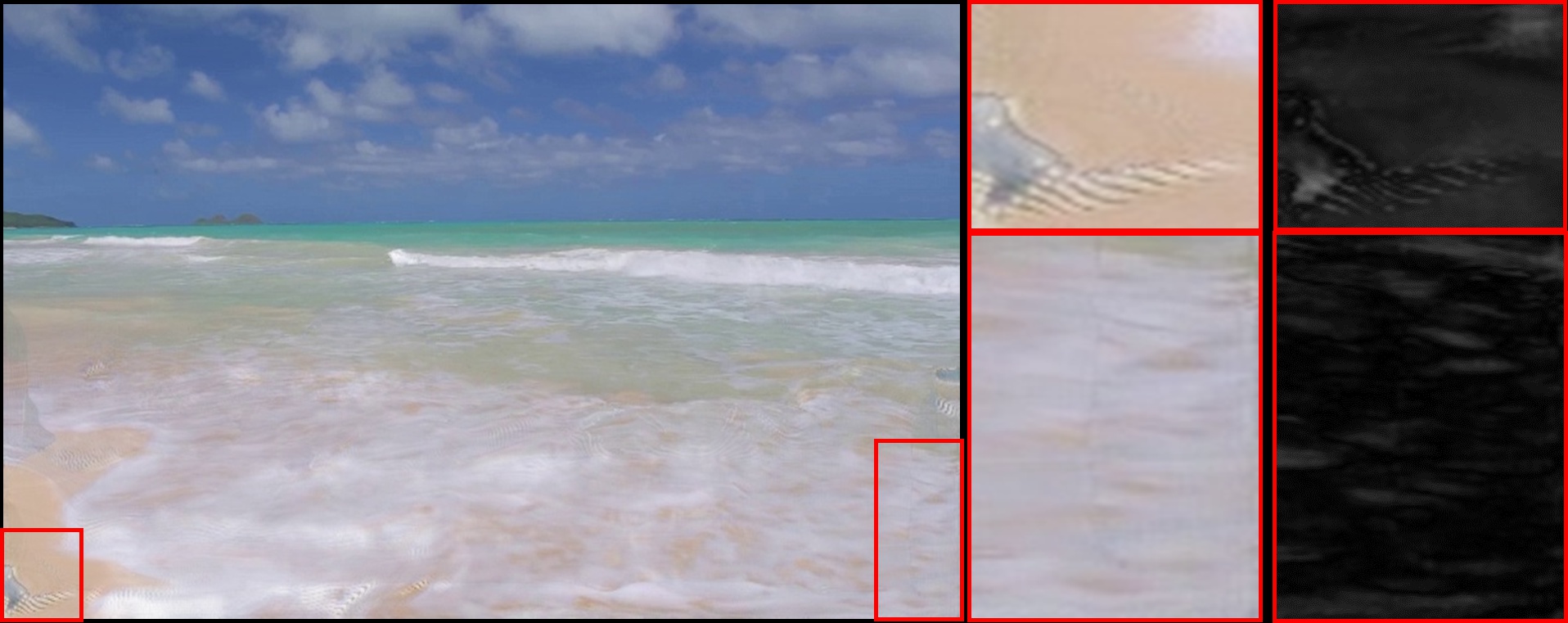}
    \includegraphics[width=\linewidth]{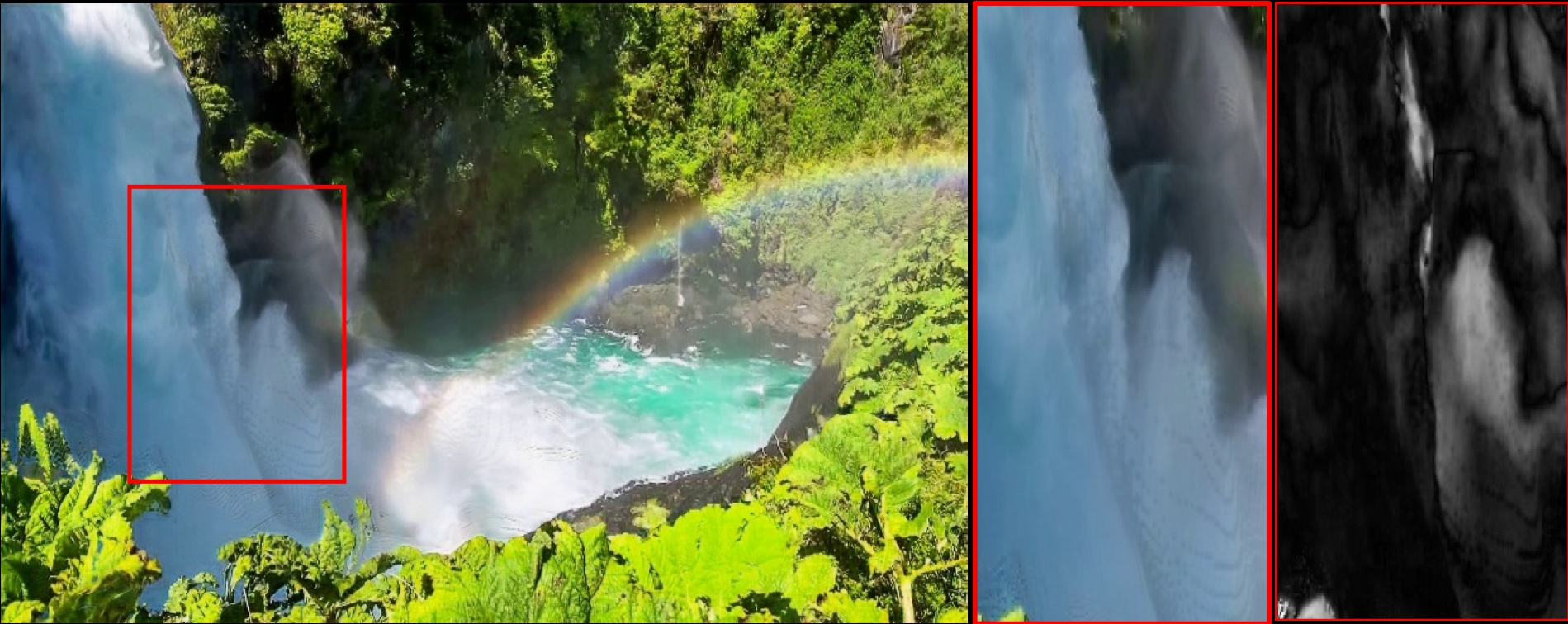}
    \caption{3D-Cinemagraphy}
    \label{fig:3dc}
  \end{subfigure}
  \begin{subfigure}{0.25\linewidth}
    \includegraphics[width=\linewidth]{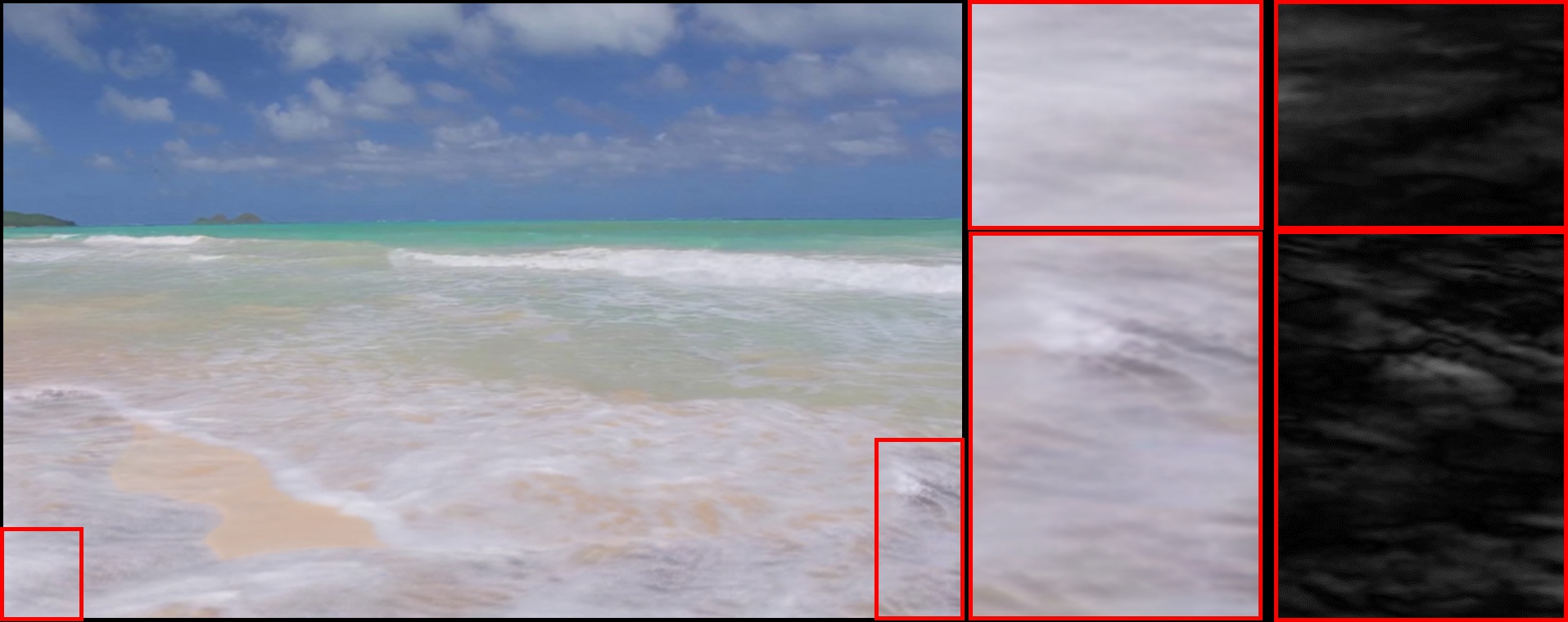}
    \includegraphics[width=\linewidth]{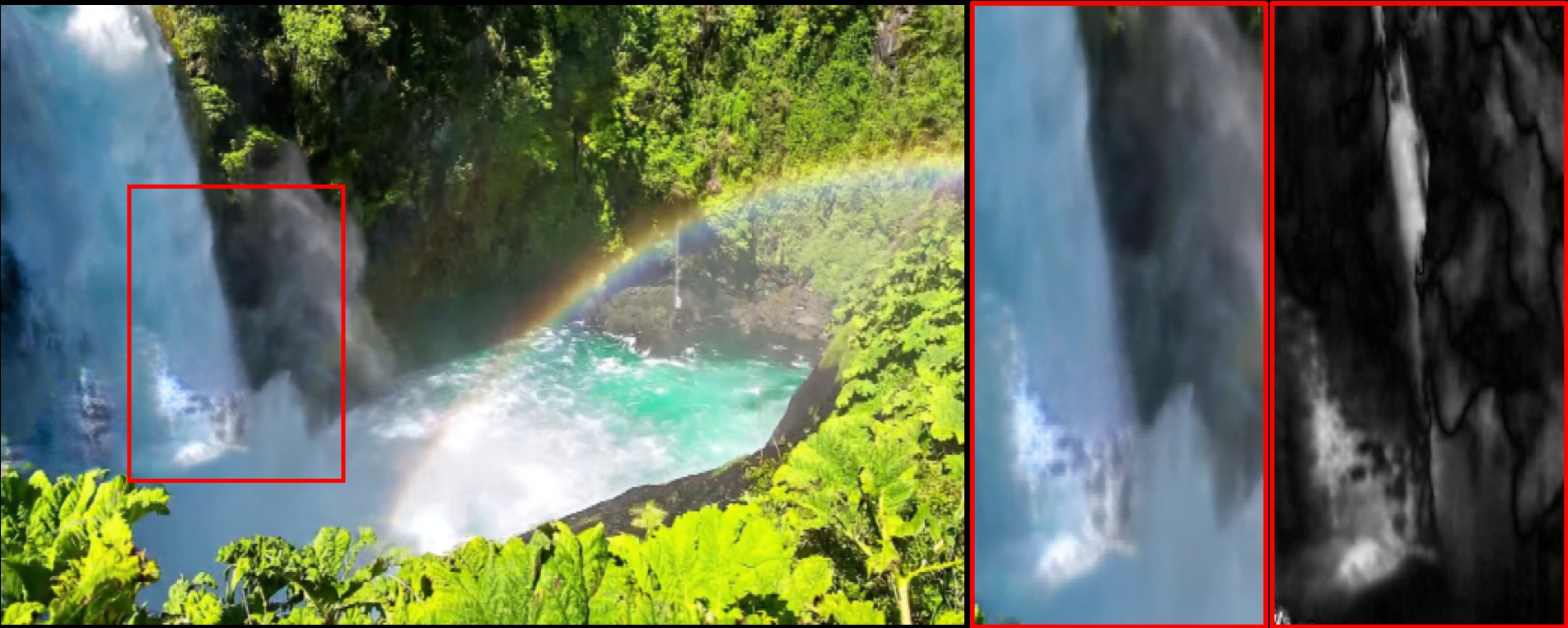}
    \caption{Holynski et al.}
    \label{fig:holy}
  \end{subfigure}
  \begin{subfigure}{0.25\linewidth}
    \includegraphics[width=\linewidth]{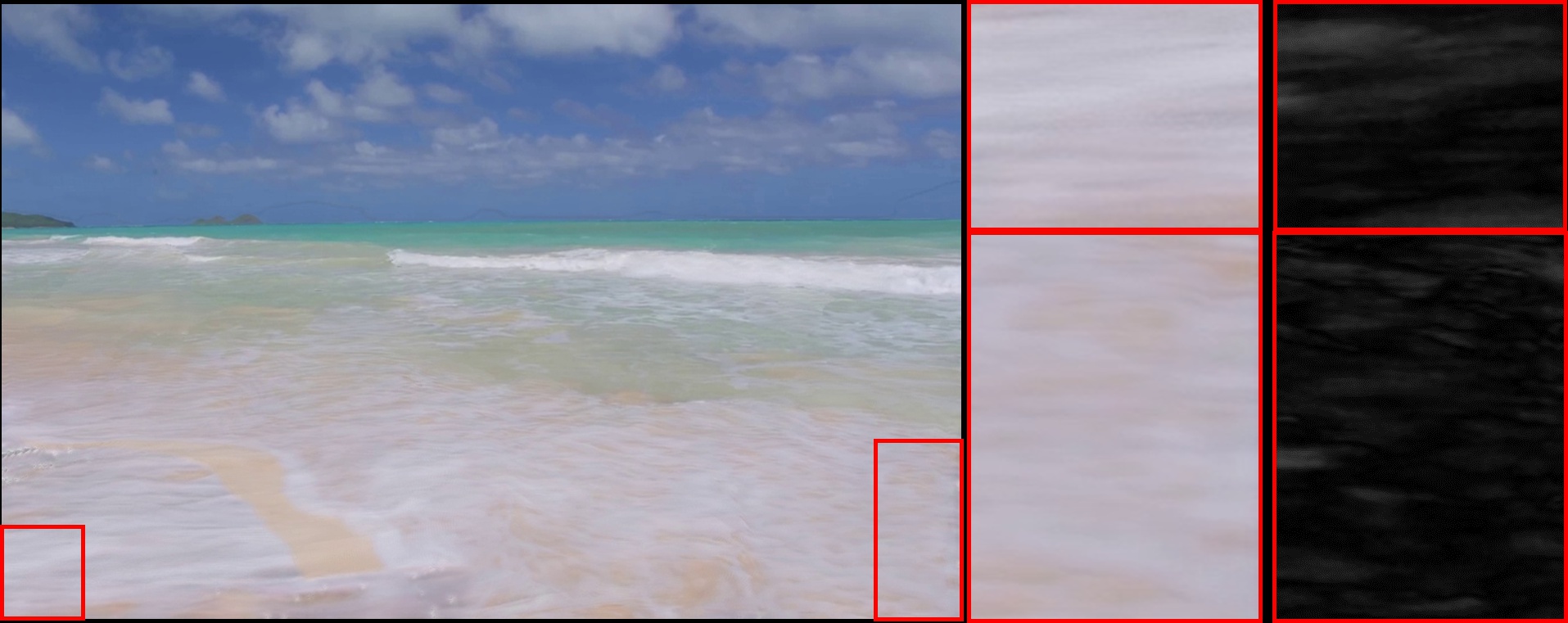}
    \includegraphics[width=\linewidth]{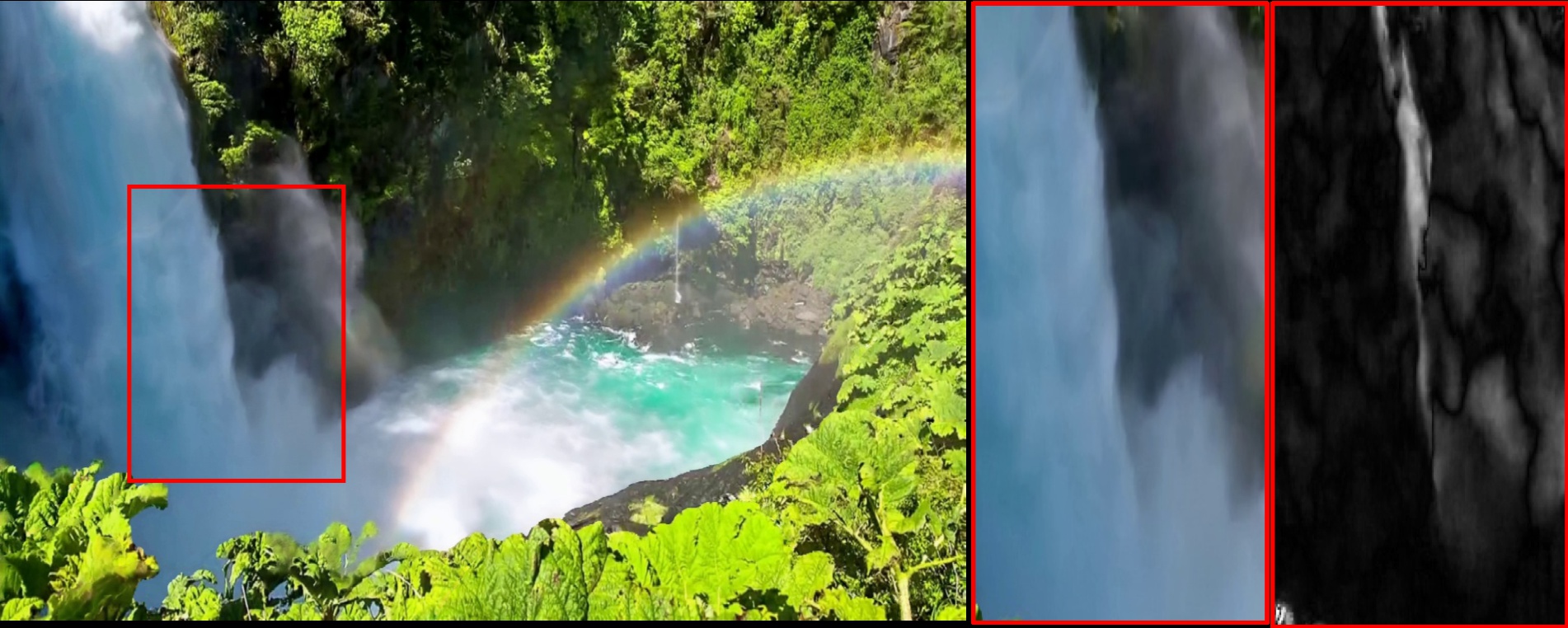}
    \caption{Ours}
    \label{fig:ours}
  \end{subfigure}
  \vspace{-6pt}
  \caption{{\bf Qualitative results on Holynski \etal~\cite{holynski}  validation set from the input view.} Our method produces compelling results, while others exhibit artifacts such as holes, dots, and mosaic patterns. Zoomed-ins and deviation maps are also provided for better visualizations. }
  \vspace{-6pt}
  \label{fig:orig}
\end{figure*}

\begin{figure*}
  \centering
  \begin{subfigure}{0.15\linewidth}
    \includegraphics[height=.65\linewidth]{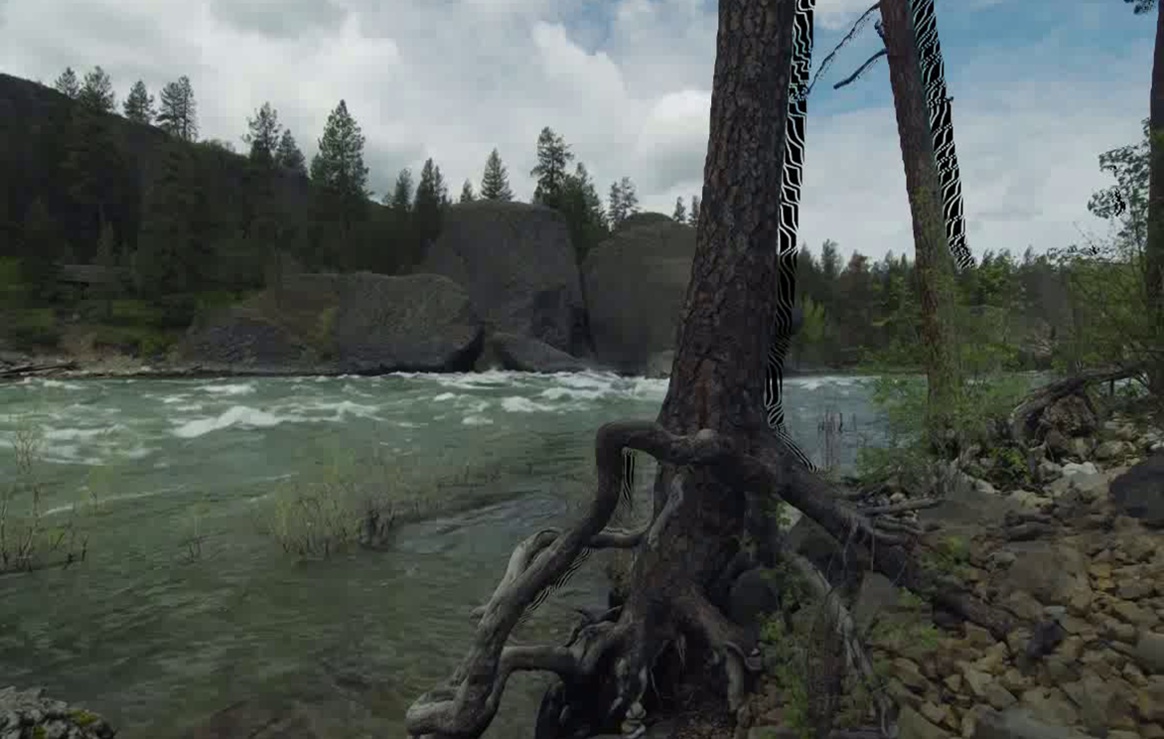}
    \includegraphics[height=.65\linewidth]{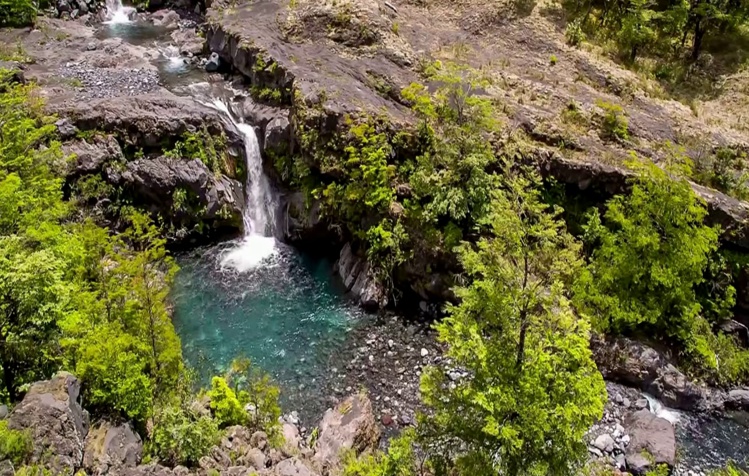}
    \includegraphics[height=.65\linewidth]{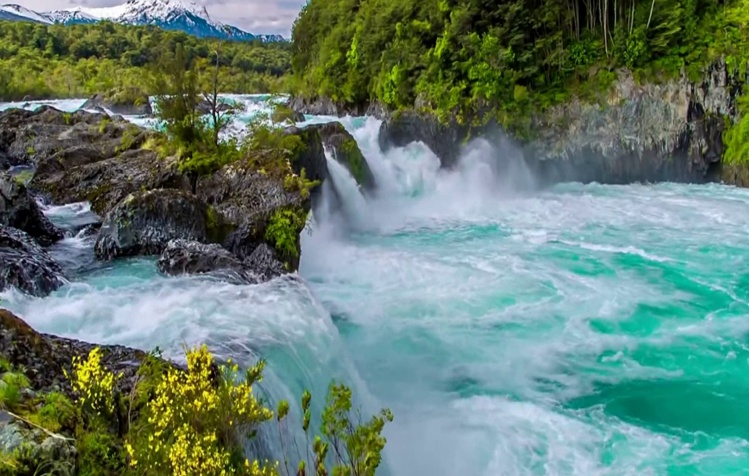}
    \caption{Ground Truth}
    \label{subfig:novel-gt}
  \end{subfigure}
  \begin{subfigure}{0.25\linewidth}
    \includegraphics[width=\linewidth]{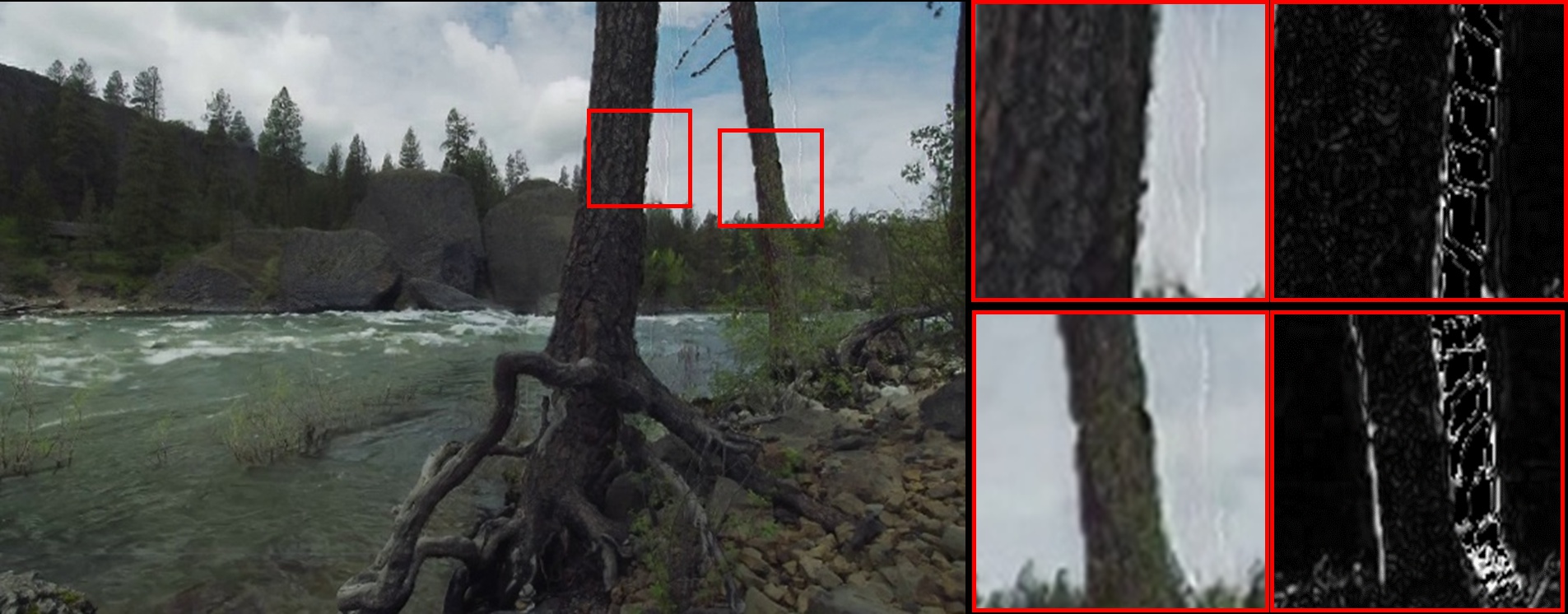}
    \includegraphics[width=\linewidth]{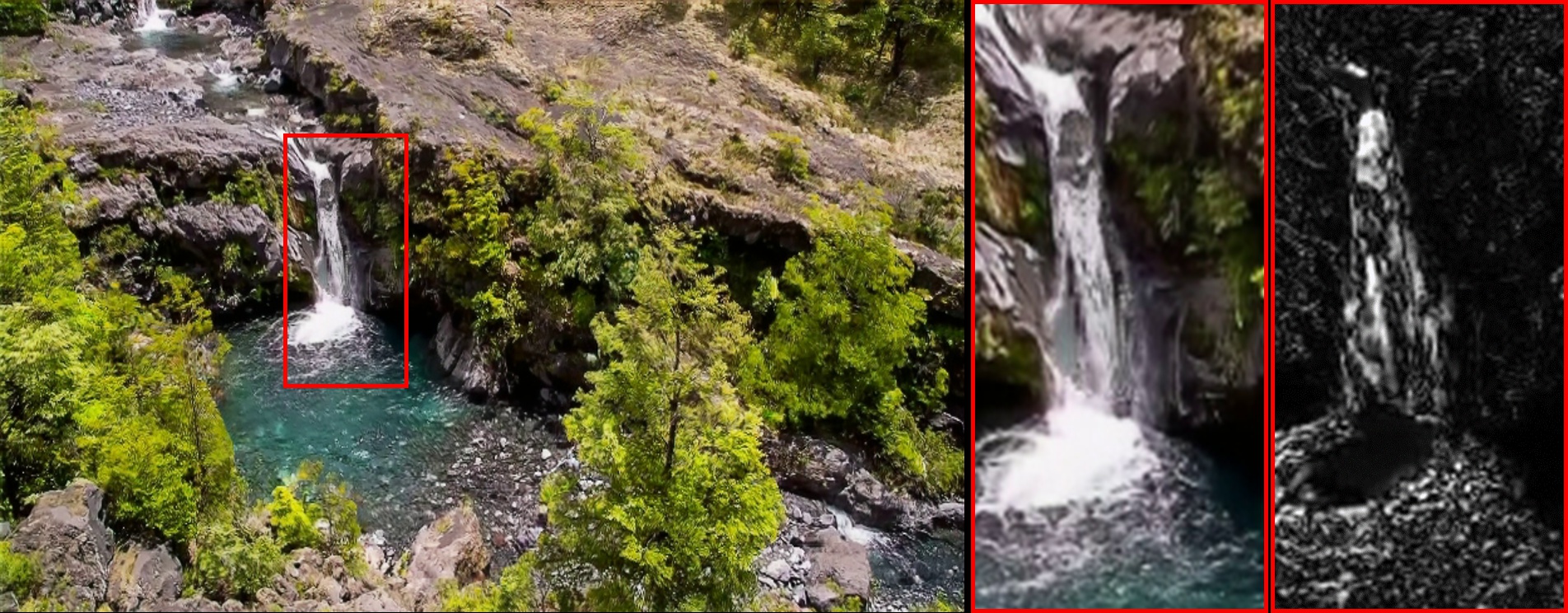}
    \includegraphics[width=\linewidth]{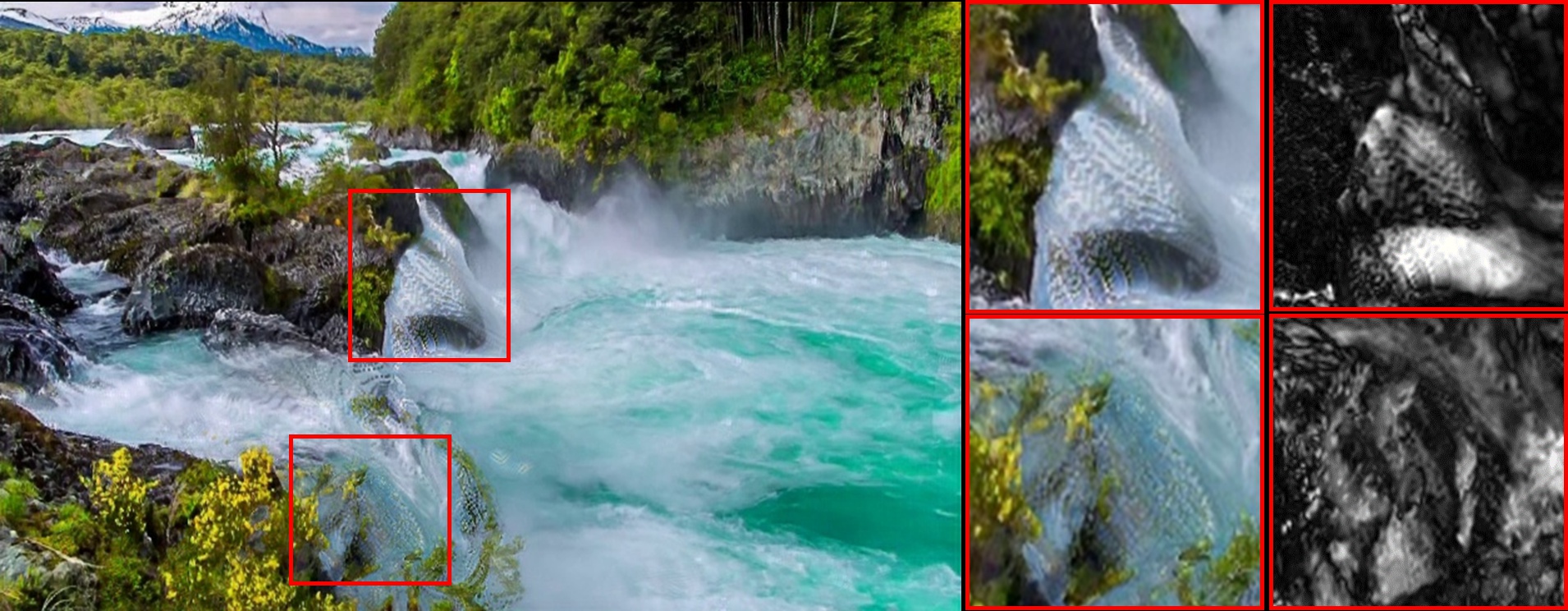}
    \caption{3D-Cinemagraphy}
    \label{fig:3dc-n}
  \end{subfigure}
  \begin{subfigure}{0.25\linewidth}
    \includegraphics[width=\linewidth]{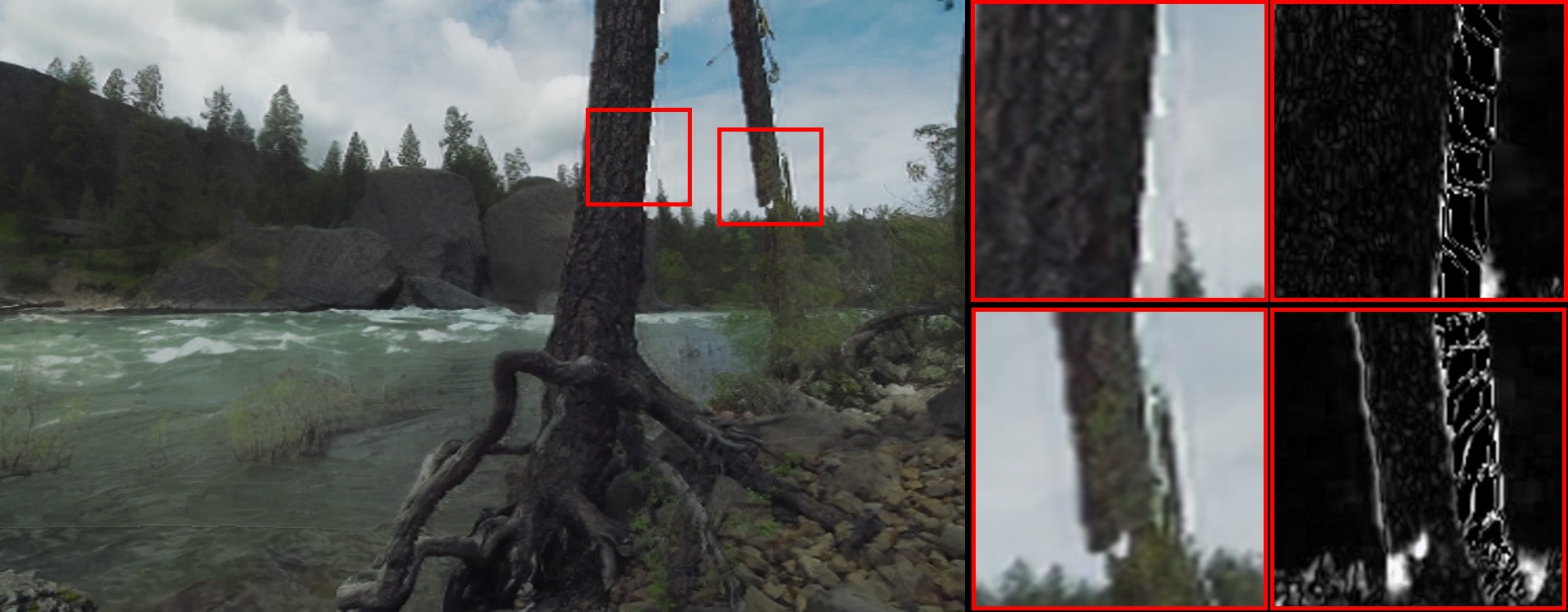}
    \includegraphics[width=\linewidth]{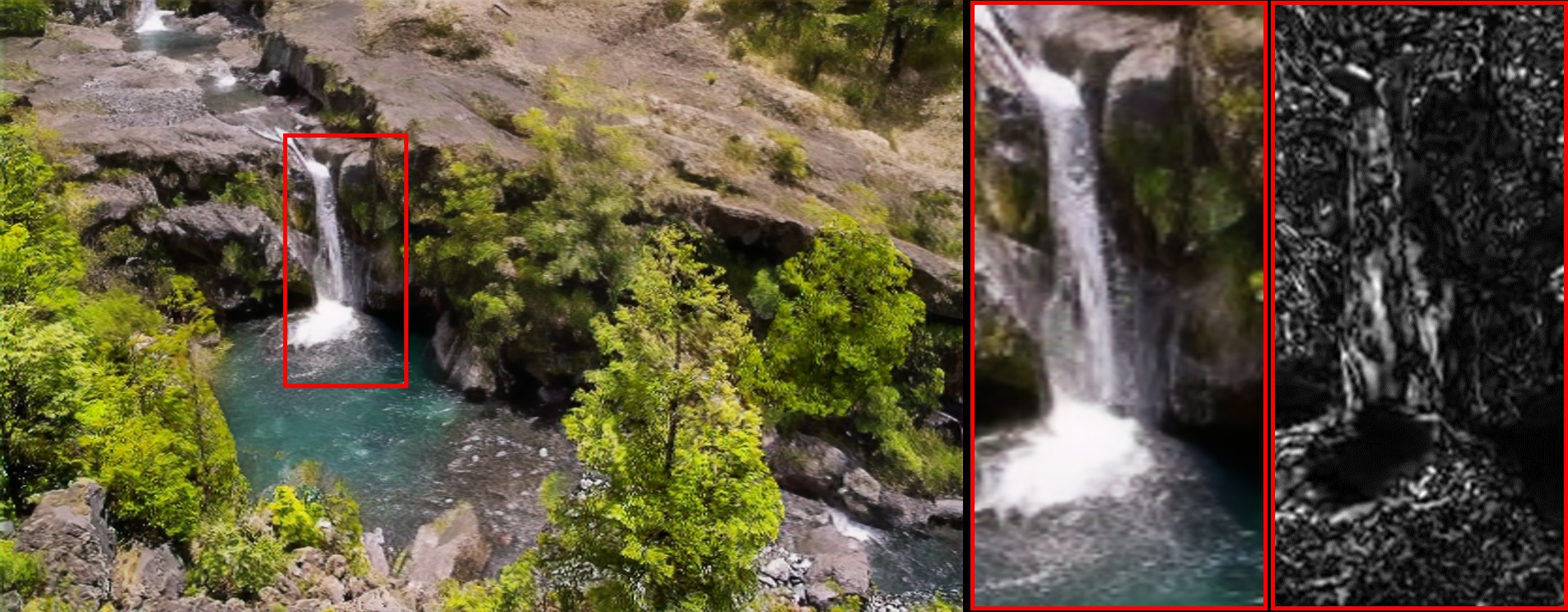}
    \includegraphics[width=\linewidth]{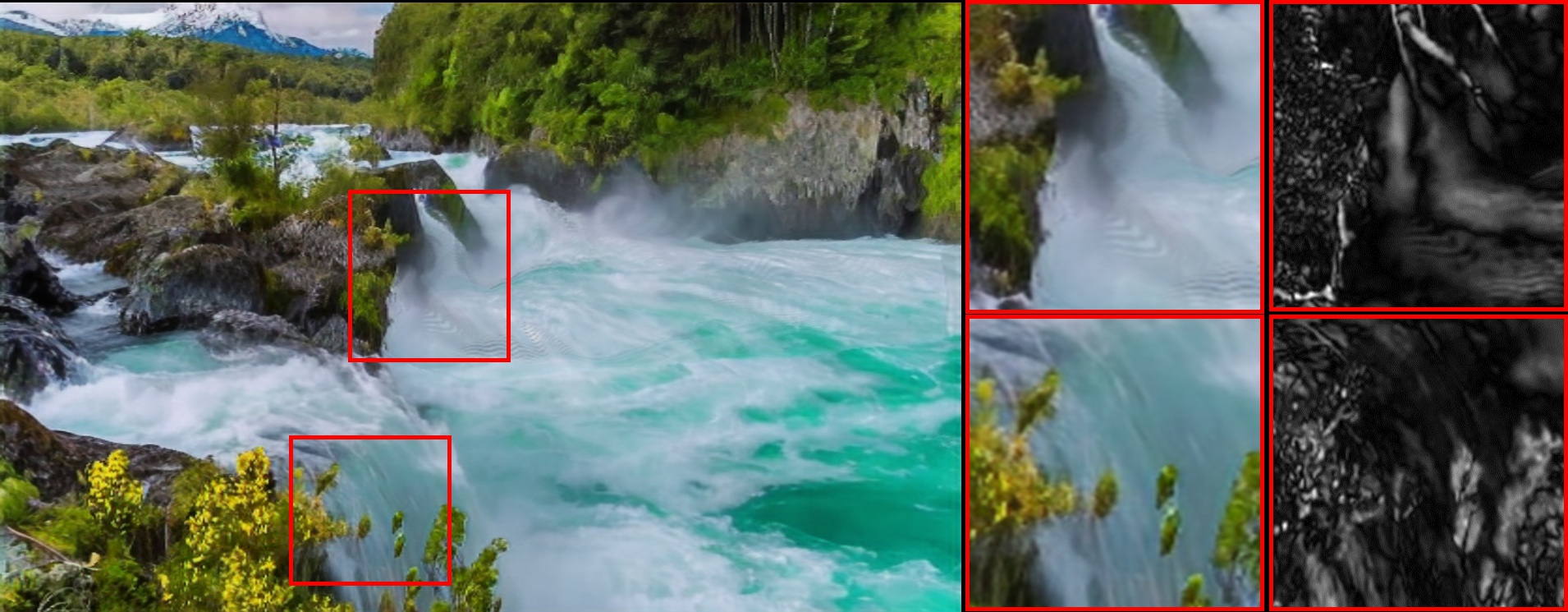}
    \caption{Make-It-4D}
    \label{fig:make-it-4d}
  \end{subfigure}
  \begin{subfigure}{0.25\linewidth}
    \includegraphics[width=\linewidth]{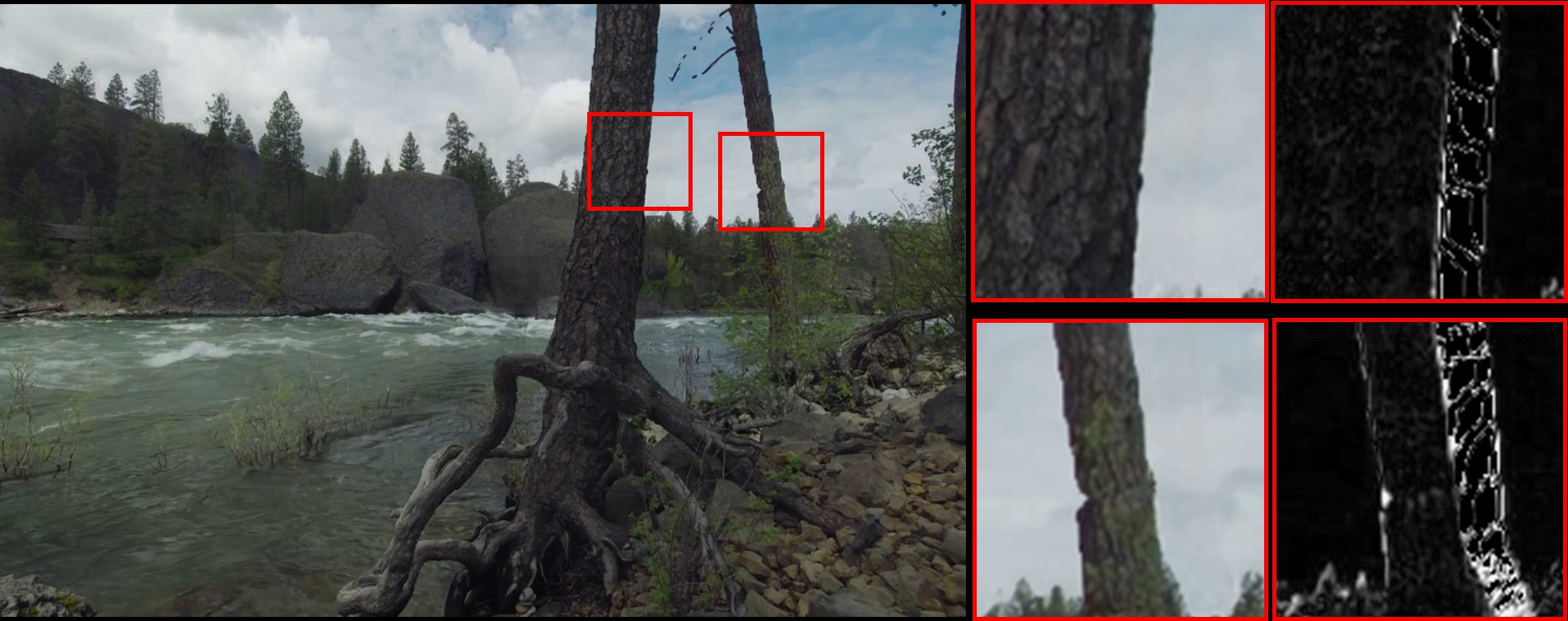}
    \includegraphics[width=\linewidth]{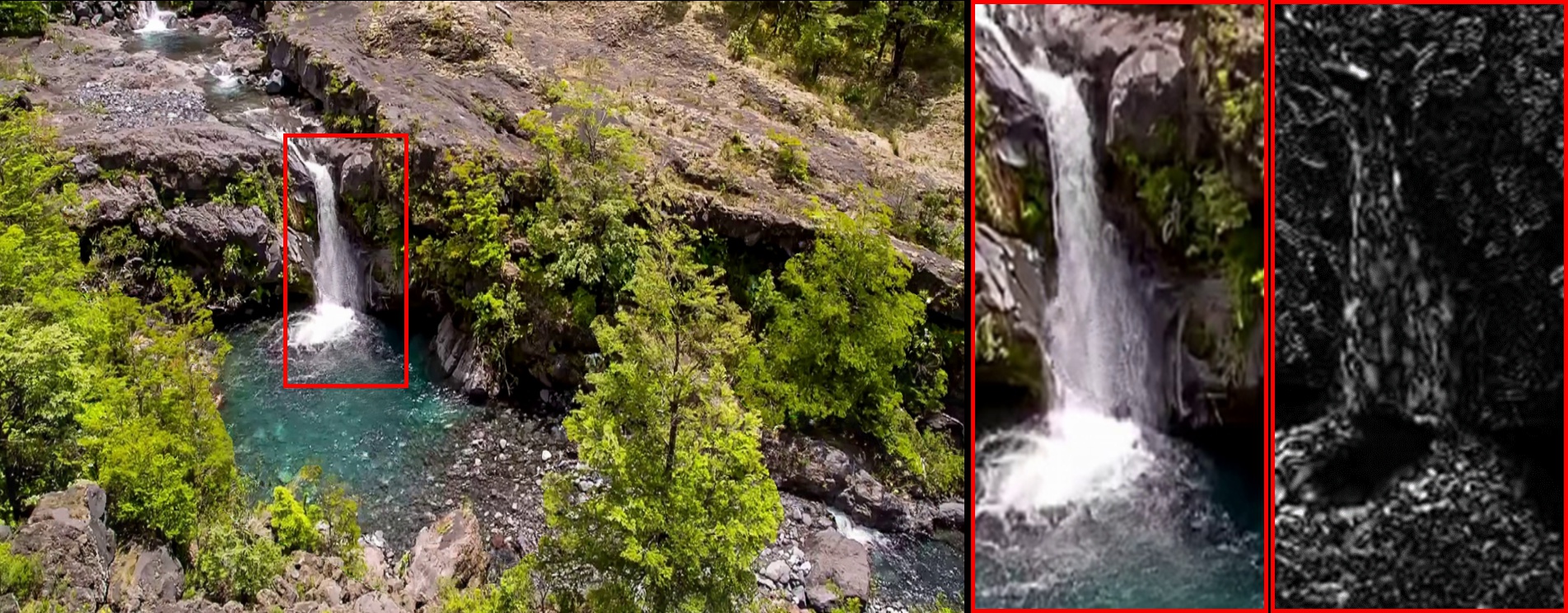}
    \includegraphics[width=\linewidth]{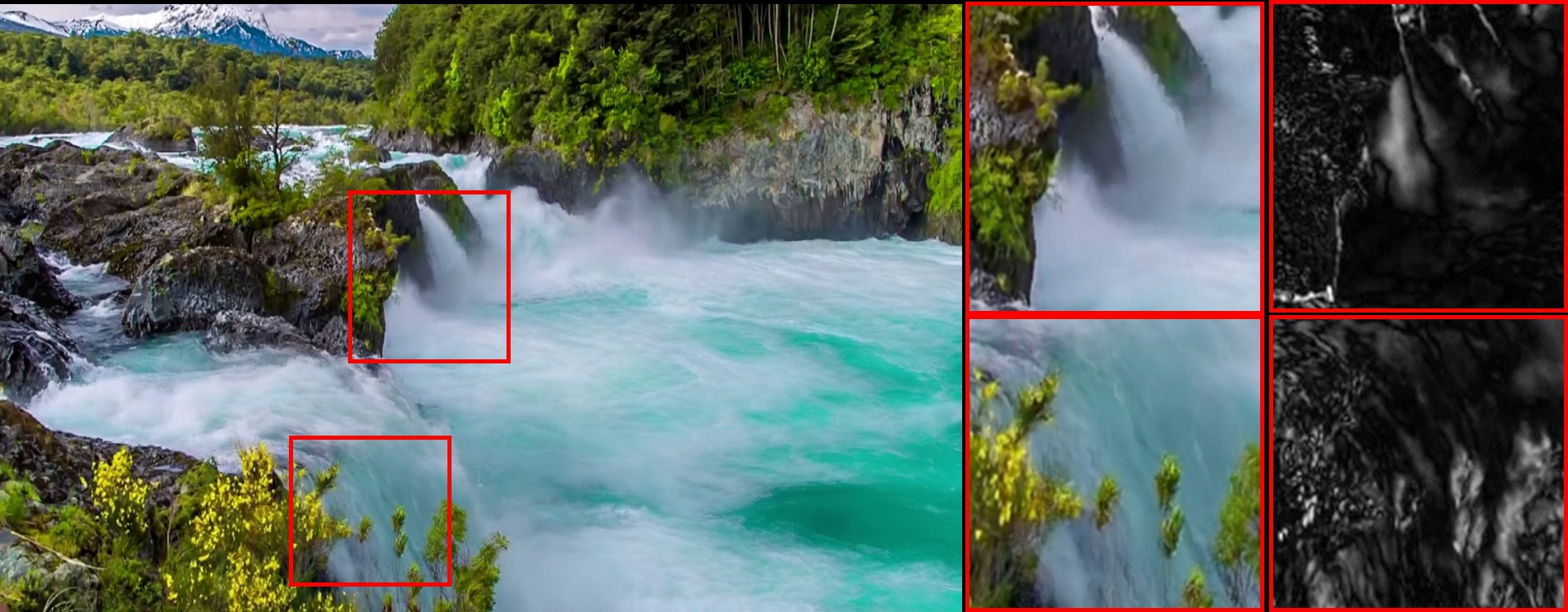}
    \caption{Ours}
    \label{fig:ours-n}
  \end{subfigure}
  \vspace{-6pt}
  \caption{{\bf Qualitative results on Holynski \etal~\cite{holynski} validation set from novel views.} Our method produces compelling results, whereas other methods exhibit the same artifacts present in videos from the input view, along with new artifacts in occluded areas. \Cref{subfig:novel-gt} shows the pseudo ground truth images obtained by rendering each frame of the ground truth videos from novel views.}
  \label{fig:novel}
  \vspace{-10pt}
\end{figure*}

\section{Experiments}
\label{sec:Experiments}

\begin{table}
    \small
    \centering
    \begin{tabular}{c|c|c}
        \toprule
        Viewpoint & Method & Preference \\
        \midrule
        \multirow{2}{*}{Input} & 3D-Cine~\cite{3d-cinemagraphy} / Ours & 30.6\% / {\bf 69.4\%}   \\
         &  Holynski \etal~\cite{holynski} / Ours & 24.5\% / {\bf 75.5\%} \\
        \hline
         \multirow{2}{*}{Novel} &  3D-Cine~\cite{3d-cinemagraphy} / Ours & 32.7\% / {\bf 67.3\%}  \\
         &  Make-it-4D~\cite{make-it-4d} / Ours & 28.3\% / {\bf 71.7\%} \\
         \bottomrule    
    \end{tabular}
    
    \vspace{-6pt}
    \caption{User study for videos from the input view and novel views. 3D-Cine refers to 3D Cinemagraphy~\cite{3d-cinemagraphy}. }
    \vspace{-15pt}
    \label{tab:user_study}
\end{table}

In this section, we present the experimental results. Our experiments aim to address the following questions: 

$\bullet$ Does our method produce better video generation results? (\Cref{subsec:quantitative}, \Cref{subsec:qualitative} and \Cref{subsec:user_study})

$\bullet$ Does physics-informed neural dynamics predict physics-bounded fluid motions? (\Cref{subsec:velocity})

$\bullet$ Does using 3D Gaussians enhance the quality of the generated videos? (\Cref{subsec:ablation})

$\bullet$ Does incorporating physics-informed neural dynamics lead to more realistic animations? (\Cref{subsec:image_editing})




\subsection{Dataset}

We train our pipeline on the training set from Holynski \etal~\cite{holynski}. We use the Holynski \etal~\cite{holynski} validation set for evaluations since the ground truth videos for the test set are not publicly available. The validation set includes 162 video clips depicting 40 distinct natural scenes. 


\begin{figure*}
  \centering
  \begin{minipage}{0.15\linewidth}
      \vspace{0.03\linewidth}
      \begin{subfigure}[b]{\linewidth}
        \centering
        \includegraphics[height=.8\linewidth, width=.8\linewidth]{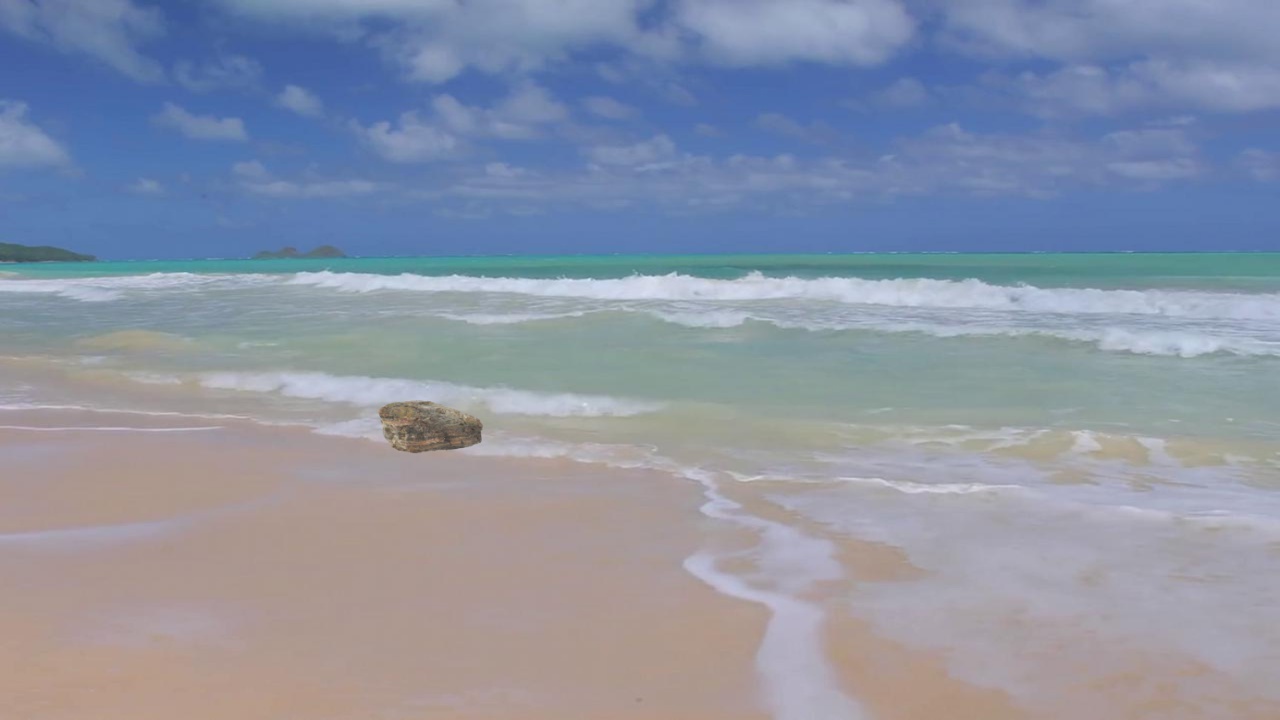}
        \vspace{0.07\linewidth}
        \caption{Edited Input Image}
        \vspace{0.03\linewidth}
      \end{subfigure}
      \begin{subfigure}[b]{\linewidth}
        \centering
        \includegraphics[height=.8\linewidth, width=.8\linewidth]{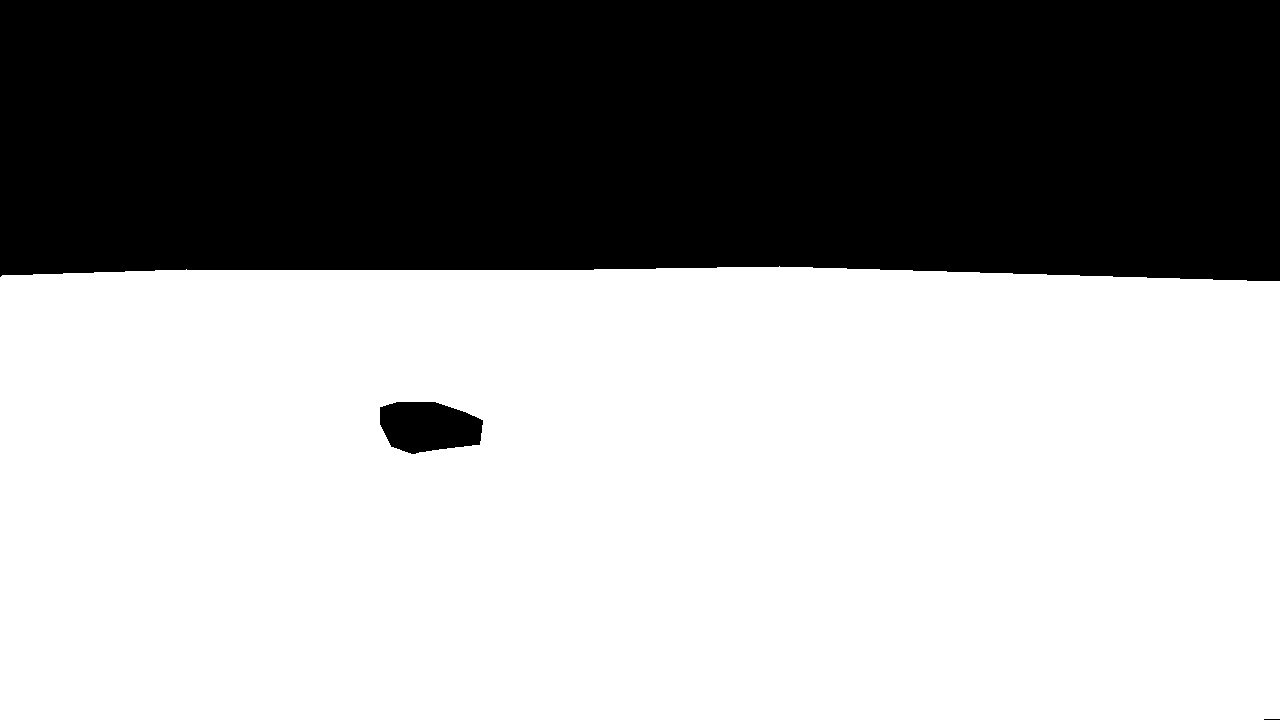}
        \vspace{0.07\linewidth}
        \caption{Edited Input Mask}
      \end{subfigure}
  \end{minipage}
  \hspace{0.02\linewidth}
  \begin{minipage}{0.8\linewidth}
      \begin{subfigure}[b]{\linewidth}
        \includegraphics[width=\linewidth]{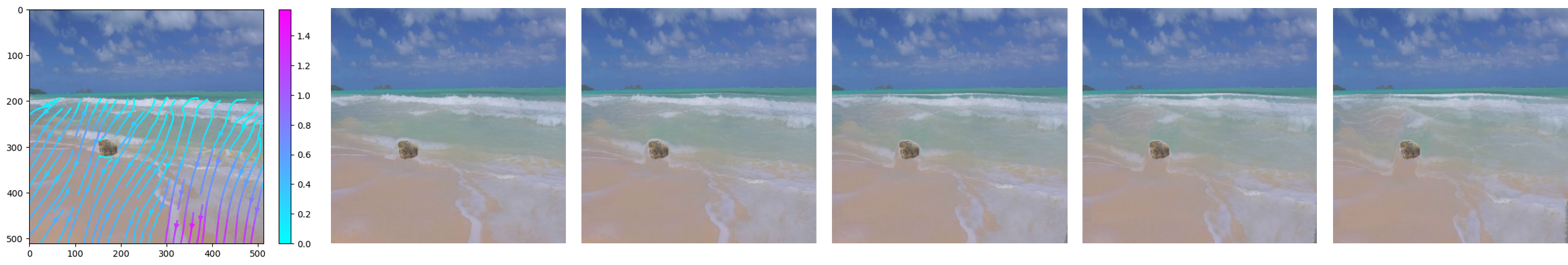}
        \caption{Streamline Plot and Key Frames - Ours}
        \label{subfing:ours}
      \end{subfigure}
      \begin{subfigure}[b]{\linewidth}
        \includegraphics[width=\linewidth]{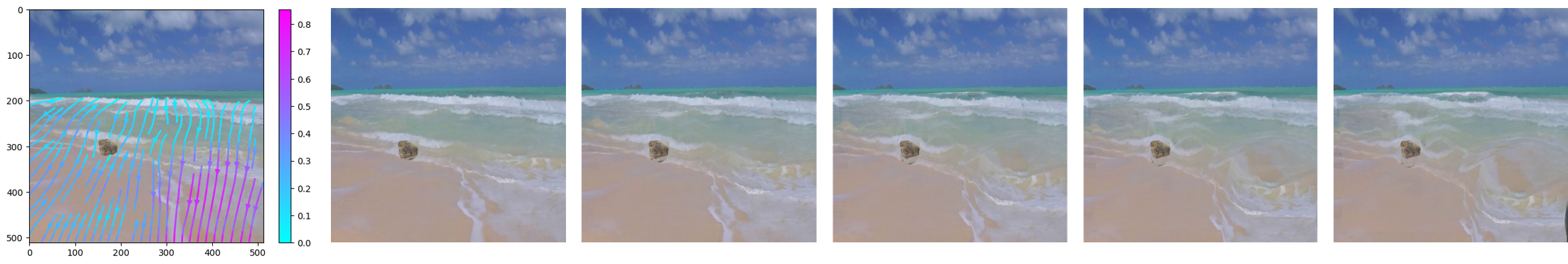}
        \caption{Streamline Plot and Key Frames - 3D-Cinemagraphy~\cite{3d-cinemagraphy}}
        \label{subfig:edit_3dc}
      \end{subfigure}
  \end{minipage}
  \vspace{-6pt}
  \caption{{\bf Visualizing Results for Image Editing}. We add an extra stone to the image and adjust the mask accordingly. We show the edited images and the corresponding streamlines predicted by our method and 3D-Cinemagraphy~\cite{3d-cinemagraphy}.}
  \label{fig:edit}
  \vspace{-15pt}
\end{figure*}

\subsection{Baselines}
Only a few works have explored 4D scene generation. 3D-Cinemagraphy~\cite{3d-cinemagraphy} is the first to tackle this task for natural fluid images, while Make-It-4D~\cite{make-it-4d} then extends 3D-Cinemagraphy~\cite{3d-cinemagraphy} to support longer camera trajectories. We treat both works as our primary baselines. 

Besides these 3D-based methods, we also compare our approach with 2D-based methods. For video generation experiments from the input view, we compare our method with Holynski \etal~\cite{holynski}, which uses a 2D flow-based animation technique to warp pixel features with predicted Eulerian motion maps. For experiments from novel camera views, we adopt the {\bf 2D Anim→NVS} and {\bf NVS→2D Anim} baselines from 3D-Cinemagraphy~\cite{3d-cinemagraphy}. As suggested by the names, both methods contain two stages, where {\bf 2D Anim} refers to generating fluid animations from a single 2D image using Holynski \etal~\cite{holynski} and {\bf NVS} refers to novel-view synthesis using 3D photo inpainting~\cite{3d-photo}. 

All methods are trained for at least 50 epochs (250k iterations) and evaluated under the setting with sparse hints. 



\begin{figure}
  \centering
  \begin{subfigure}{0.4\linewidth}
      \includegraphics[width=\linewidth]{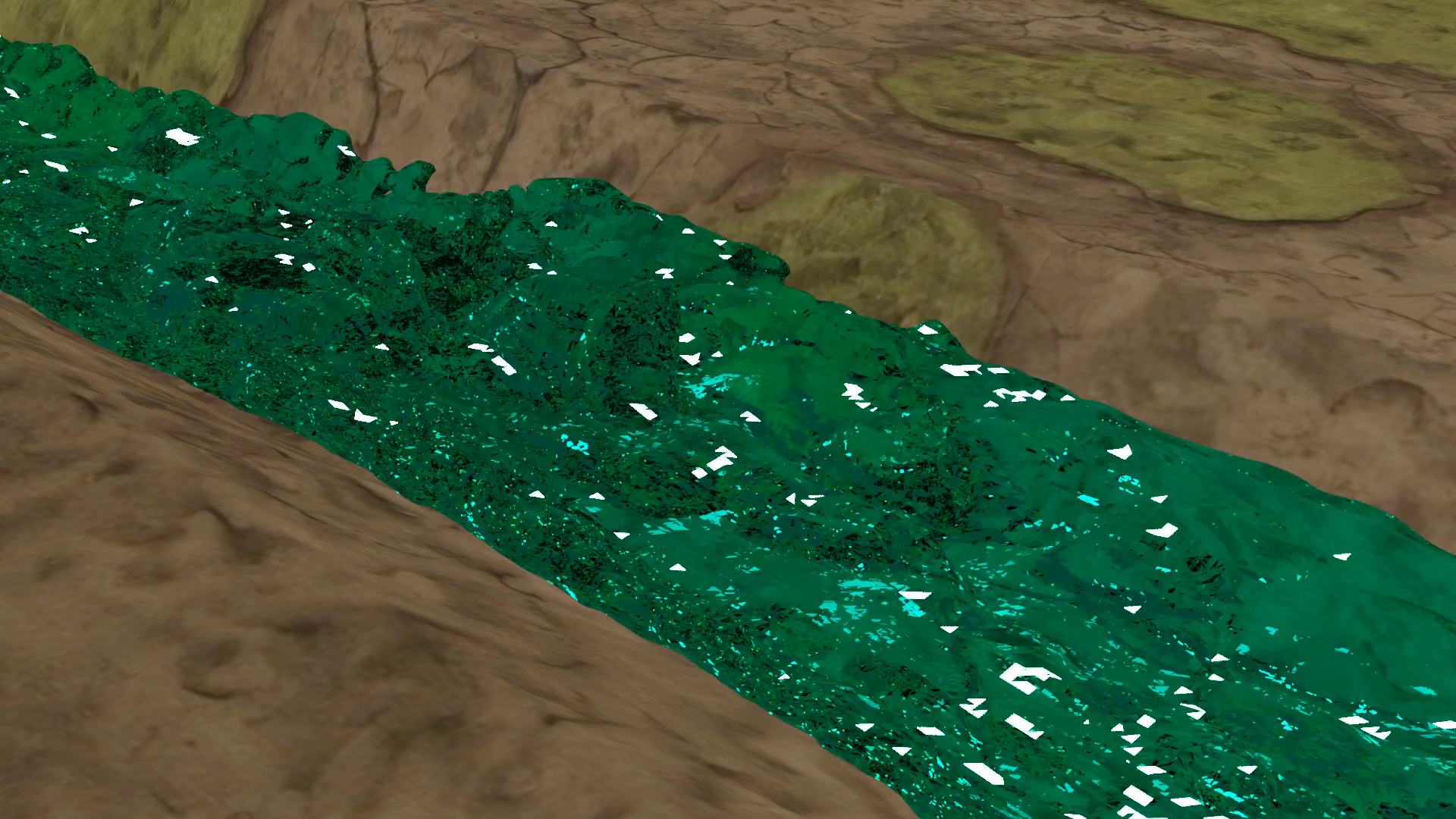}
      \caption{Without the rock}
  \end{subfigure}
  \begin{subfigure}{0.4\linewidth}
    \includegraphics[width=\linewidth]{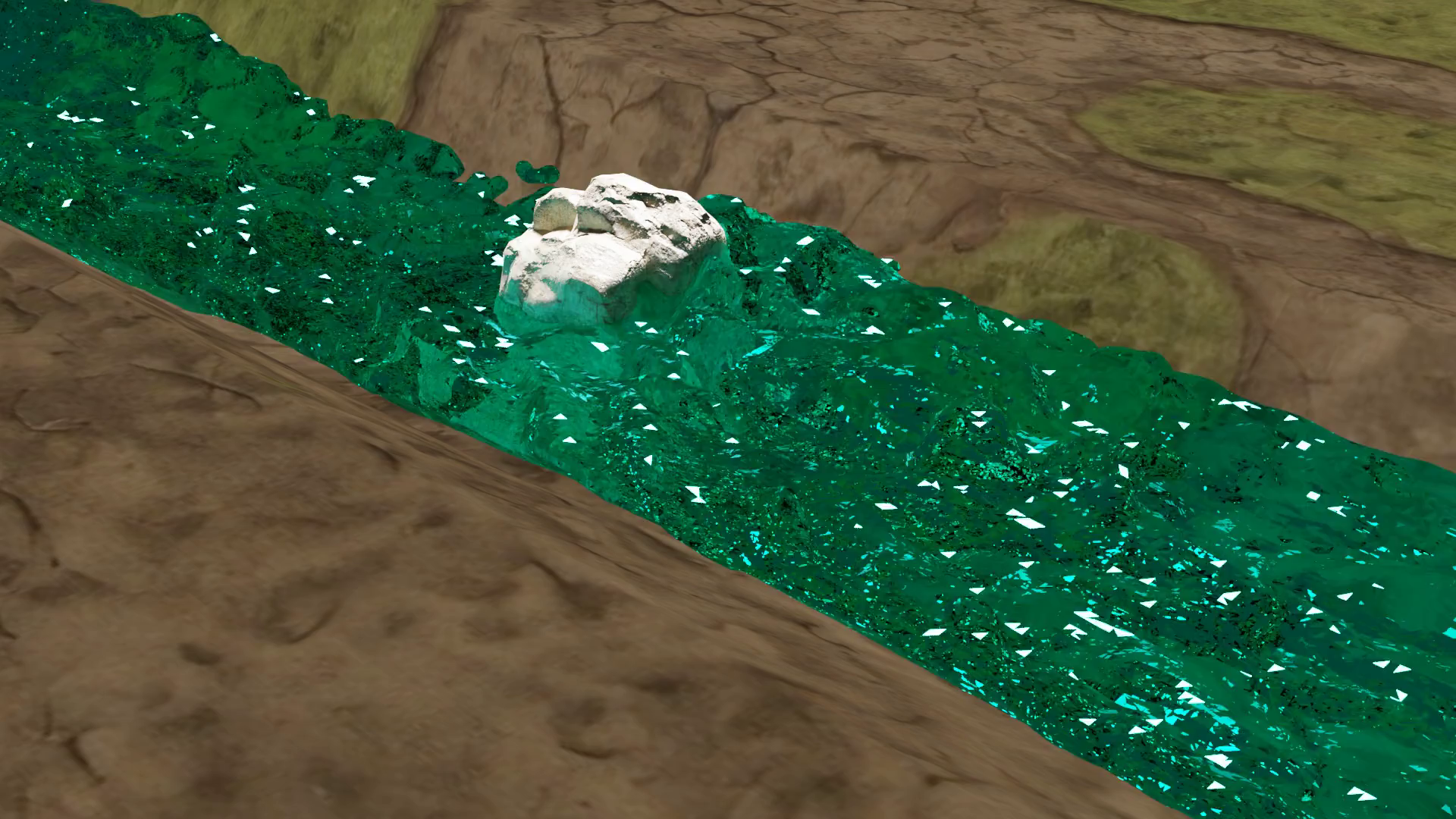}
    \caption{With the rock}
  \end{subfigure}
  \vspace{-8pt}
  \caption{A visualization of our synthetic scenes.}
  \label{fig:toy_scenes}
  \vspace{-10pt}
\end{figure}

\subsection{Quantitative Comparisons}
\label{subsec:quantitative}
We perform two experiments on the Holynski \etal~\cite{holynski} validation set: one generates videos from the viewpoint of the input image, the other from novel views. The results are shown in \Cref{tab:orig} and \Cref{tab:sep_novel}, respectively. For the novel-view experiment, we generate pseudo ground truth by rendering incomplete novel views per frame, following ~\cite{3d-cinemagraphy}. Metrics are computed only in regions visible in the input images. We report image‑level metrics (PSNR, SSIM, LPIPS~\cite{lpips}) and the video‑level metric VMAF.

As shown in \Cref{tab:orig} and \Cref{tab:sep_novel}, our method significantly exceeds baselines in all metrics, improving PSNR from 22.81 to 24.98 for videos from the input view and from 22.46 to 24.34 for videos from the novel view. This improvement is attributed to: (a) our physics-informed neural dynamics, which produce more realistic fluid motions, and (b) 3D Gaussians, which enhance scene appearance reconstruction. To further analyze the contributions of (a) and (b) separately, we report results for fluid and background regions independently. Our method achieves noticeable improvements in both regions.

\subsection{Qualitative Comparisons}
\label{subsec:qualitative}
\Cref{fig:orig} compares the methods on the Holynski \etal~\cite{holynski} validations set from the input view. Our method achieves better visual quality. 3D-Cinemagraphy~\cite{3d-cinemagraphy} shows artifacts such as holes (row 1) and dots in fluids (row 2). Holynski \etal~\cite{holynski} introduces mosaic-patterned artifacts. Artifacts in both methods occur in fast-moving regions, where holes form as fluids move forward. Despite using symmetric splatting, our baselines struggle due to the limitation of their fluid representations. In contrast, our 3D Gaussian representation naturally blends and smooths hole regions.



\Cref{fig:novel} shows the results from novel views. The aforementioned artifacts still exist in the novel-view videos generated by our baselines. Besides, our method better reconstructs the regions invisible from the input view (first row), thanks to 3D Gaussian representations. In comparison, 3D-Cinemagraphy~\cite{3d-cinemagraphy} blurs in invisible regions and Make-It-4D~\cite{make-it-4d} introduces redundant structures. 


\subsection{User Study}
\label{subsec:user_study}

We conduct a user study with 30 participants to assess our method from a human perspective. We select 40 images from the Holynski \etal~\cite{holynski} set, using 20 for the input view and 20 for novel view generation. Participants are asked to compare video pairs generated from the same image and select the higher-quality one. The results of the user study are shown in \Cref{tab:user_study}, where our baselines achieve an average improvement of 42\% in human preference.



\subsection{Image Editing}
\label{subsec:image_editing}
Our model follows physical laws, enabling editing boundary conditions in the input image while generating plausible results. \Cref{fig:edit} shows an example where a rock is added to obstruct water flow. Unlike SLR-SFS~\cite{SLR-SFS}, our method only requires modifying the input image and fluid mask, without the need for a 3D model. We compare the results from our method and 3D-Cinemagraphy~\cite{3d-cinemagraphy} in \Cref{fig:edit}.

As shown in \Cref{fig:edit}, the motions predicted by 3D-Cinemagraphy\cite{3d-cinemagraphy} penetrate the occlusion, as their motion predictor relies solely on data priors. In contrast, our method, bounded by physical laws, generates more plausible results with fluids properly separated by the rock.

\subsection{Quantitative Analysis of Velocity Prediction}
\label{subsec:velocity}
In the previous sections, we show that physics-informed neural dynamics generates more realistic animations and improves video quality. To validate the accuracy of our predicted velocity fields, we construct two synthetic scenes with water flows and provide quantitative results for velocity prediction. Shown in \Cref{fig:toy_scenes}, the two scenes mimic natural rivers, with one featuring a rock to obstruct the flow.

For both scenes, we compute the velocities of surface points and report L1 errors. Shown in \Cref{tab:synthetic}, our method significantly outperforms 3D-Cinemagraphy~\cite{3d-cinemagraphy}, achieving roughly 22\% improvement in the scene with the rock.  

\begin{table}
  \small
  \centering
   \begin{tabular}{cc | lll}
        \toprule
        Physics & Gaussians & PSNR↑ & SSIM↑ & LPIPS↓ \\
        \midrule
          &    & 21.49  & 0.65  & 0.25  \\
         & \checkmark & 22.58 & 0.70 & {\bf 0.22} \\
        \checkmark & \checkmark  & {\bf 23.05} & {\bf 0.72} & 0.26 \\
        \bottomrule
    \end{tabular}
    \vspace{-6pt}
    \caption{Ablation study of 3D Gaussians and physics-informed neural dynamics on Holynski \etal \cite{holynski} validation set. }
    \label{tab:ablation}
    \vspace{-10pt}
\end{table}

\begin{table}
  \centering
  \small
  \begin{tabular}{@{}lclll}
    \toprule
     &  w/o rock & w/ rock \\
    \midrule
    Ours & \textbf{0.0389} & \textbf{0.0388} \\
    3D-Cinemagraphy~\cite{3d-cinemagraphy} & 0.0432 & 0.0498 \\
    \bottomrule
  \end{tabular}
  \vspace{-6pt}
  \caption{L1 error for velocity prediction on our synthetic scenes. }
  \label{tab:synthetic}
  \vspace{-10pt}
\end{table}

\begin{table}
   \centering
   \small
   \begin{tabular}{l | ccc}
        \toprule
         & w/o rock & w/ rock\\
        \midrule
        w/ viscosity & 0.0415 & 0.0501  \\
        w/ pressure fields & 0.0516 & 0.415 \\
        w/o external forces & 0.0612 & 0.0549 \\
        ours & \textbf{0.0389} & \textbf{0.0388} \\
        \bottomrule
    \end{tabular}
    \vspace{-6pt}
    \caption{Ablation study for physics-informed neural dynamics on our synthetic scenes.}
    \vspace{-15pt}
    \label{tab:ablation_physics}
\end{table}

\subsection{Ablations}
\label{subsec:ablation}
We perform two ablation studies to validate our design choices. The first study evaluates the impact of the physics-informed neural dynamics and the 3D Gaussian representation, respectively. We train both stages for 25k iterations. Shown in \Cref{tab:ablation}, using 3D Gaussians alone improves PSNR from 21.49 to 22.58, while incorporating physics-informed neural dynamics further boosts it to 23.05.

The second study validates our design choice for the physics-informed neural dynamics, in which we simplify the \Cref{eq:ns_eq} when deriving the physics guidance for the networks. As shown in \Cref{tab:ablation_physics}, our method also defeats all the alternatives. Considering viscosity requires computing a second-order derivative, increasing the optimization complexity. Adding pressure fields requires an extra network, complicating optimization and reducing performance. External forces, such as gravity, are common in our settings, so ignoring them results in unrealistic fluid motions.

\section{Conclusion}
We present a novel method for 4D scene generation from a single fluid image. Our method combines a physics-informed neural dynamics model for estimating 3D velocity fields with a feature-based 3D Gaussian representation, ensuring both photorealistic results and adherence to physical principles. Experiments show that our method outperforms existing methods, achieving physics-bounded fluid animations with higher video quality.



\section{Acknowledgements}
The USC Geometry, Vision, and Learning Lab acknowledges generous supports from Toyota Research Institute, Dolby, Google DeepMind, Capital One, Nvidia, and Qualcomm. Yue Wang is also supported by a Powell Research Award.

{
    \small
    \bibliographystyle{ieeenat_fullname}
    \bibliography{main}
}
\clearpage
\setcounter{page}{1}
\maketitlesupplementary

\begin{figure*}[ht]
  \centering
  \includegraphics[width=\linewidth]{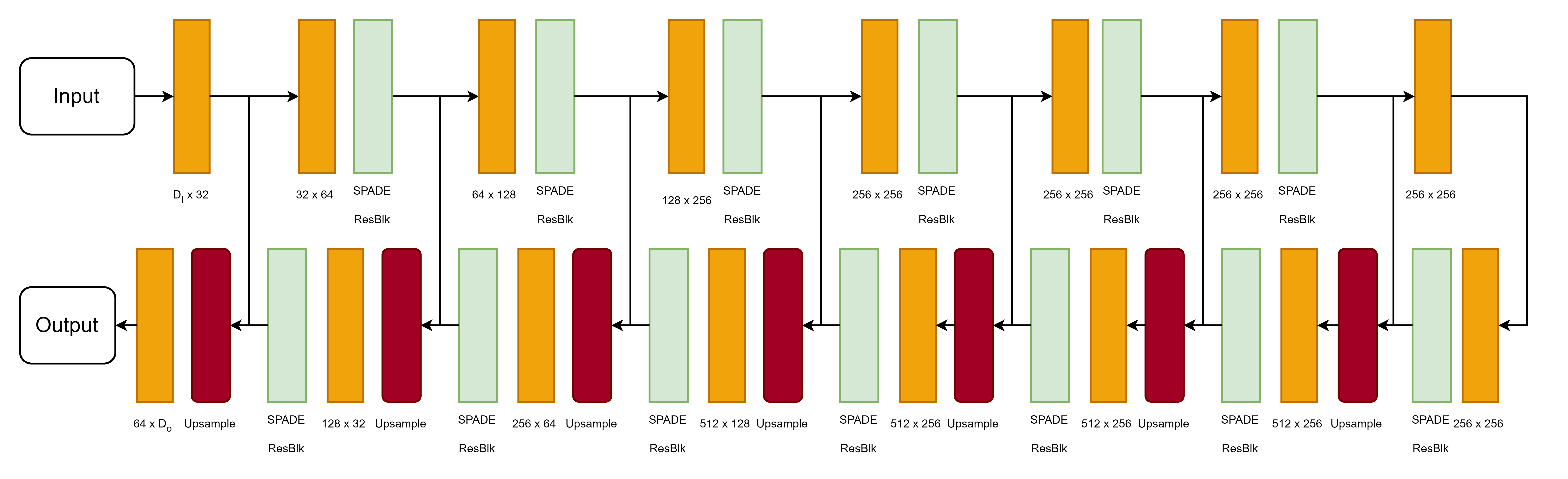}
  \caption{{\bf Detailed Architecture of our feature network.} Our feature network is designed based on SPADE \cite{spade}. The encoder is shown at the top of the figure, while the decoder is displayed at the bottom. Leaky ReLU activation is applied after each convolution layer in the encoder (indicated by orange rectangles), whereas ReLU activation is used after each convolution layer in the decoder. }
  \label{fig:feature_network}
\end{figure*}

\section{Implementation Details}
Our training process consists of two stages: first, the physics-informed neural dynamics model is trained for 120k iterations, followed by the animation module for 250k iterations. Both stages use the Adam optimizer with a 1e-4 learning rate and (0, 0.9) for betas. Training is performed on Nvidia RTX 6000 Ada GPUs with a batch size of 1. Weight decay is applied every 50k iterations with a decay rate of 0.5. 

As illustrated in the main paper, depth maps of the input image and fluid-area masks are also needed. We use ~\cite{vit} for depth prediction. Fluid area masks are manually specified by users.

For quantitative and qualitative comparisons, we follow this full training setup. However, in the first ablation study, both stages are trained for only 25k iterations on a training subset that is 10× smaller than the original dataset. Evaluations are still conducted on the full validation set. In the second ablation study, all the alternatives are trained for 120k iterations, as in the full training setup.

\section{User Hints}
\begin{figure}
  \centering
  \includegraphics[width=0.9\linewidth]{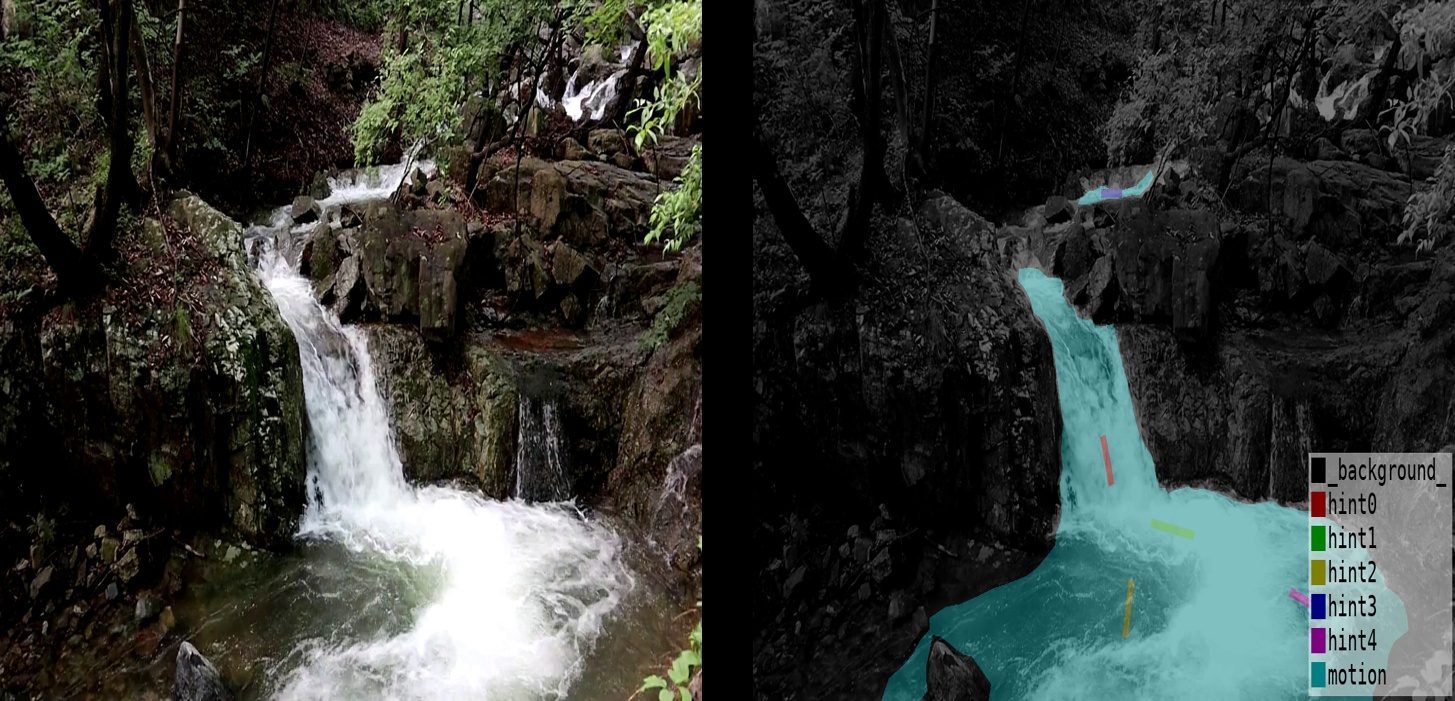}
  \caption{An example of user hints. }
  \label{fig:user_hint}
\end{figure}

Our network can be extended to incorporate sparse flow hints from users, as shown in \Cref{fig:user_hint}. These hints provide valuable information, such as the general direction of fluid flow, and help to reduce ambiguity in velocity prediction (e.g., a river could flow both leftward or rightward). Therefore, we append them to our evaluations for all the methods. 

In the user hint setting, we first convert the sparse hints into a dense hint map, which is then appended to the inputs of our physics-informed neural dynamics.

\section{Model Architecture}
Our model consists of two parts: physics-informed neural dynamics and the animation part. The animation part contains an inpainter to inpaint each LDI, an encoder to extract features for each LDI, and a decoder to decode the feature maps rendered by 3D Gaussians. Similar to 3D-Cinemagraphy~\cite{3d-cinemagraphy}, we use the inpainter from ~\cite{3d-photo}. The feature encoder mostly follows the architecture of ResNet34, while the decoder adopts the U-Net architecture. 

For our physics-informed neural dynamics, the architecture will be detailed in the following sections. The model comprises three components: a convolutional network that extracts features from the inputs, an MLP that processes the 4D coordinates along with the extracted features to predict the velocity of the points, and an additional network that predicts the external forces for each image. We refer to these three parts as Feature Network, Velocity Network, and External Force Network, respectively. 

\subsection{Feature Network}
Our feature network processes the source image, monocular depth map, and fluid area mask to generate a feature map. In the scenarios involving user interaction, the network also incorporates a user hint mask as an additional input. \Cref{fig:feature_network} illustrates the architecture of our feature network, which is based on the design of SPADE~\cite{spade}. We find the residual connection blocks in SPADE very useful as they allow the network to retain input information while processing it incrementally. In the encoder section, shown at the top of \Cref{fig:feature_network}, we use Leaky ReLU activation with a negative slope of 0.2. We set the kernel size to 4 and the stride to 2 for all the convolution layers in the encoder. For the decoder section, shown at the bottom of the figure, ReLU activation is applied. We set the kernel size to 3 and the stride to 1 for convolution layers in the decoder section. We set padding to 1 for all the convolution layers in the feature network. Additionally, we downsample all inputs to a resolution of (512, 512) for faster processing.

\subsection{Velocity Network}
Our velocity network is composed of five fully connected layers with ReLU activation applied after each layer except the final one. The dimensions of the intermediate layers are sequentially set to 2048, 1024, 256, and 64. To prevent the prediction of overly smooth velocity fields, we apply positional embedding~\cite{nerf} to the input 4D coordinates $(x,y,z,t)$. To ensure all coordinates fall within the range $[-1, 1]$, we first transform the 3D spatial locations $(x,y,z)$ using the following equations: 
\begin{equation}
    (x^p, y^p, d) = Project((x,y,z), P),
\end{equation}
where $Project(\cdot)$ indicates projecting the world coordinates to the camera plane and $P$ is the projection matrix of the input view. We then re-scale and shift $(x^p, y^p, d)$ by:
\begin{equation}
    x' = \frac{2x^p }{W-1} - 1,
\end{equation}
\begin{equation}
     y' = \frac{2y^p}{H-1} - 1,
\end{equation}
\begin{equation}
    z' = \frac{2}{\max(d, 1)} - 1,
\end{equation}
where $H$ and $W$ are the height and width of the input image. For the time $t$ in the 4D coordinates, we transform it by:
\begin{equation}
    t' = \frac{t}{T}, 
\end{equation}
where $T$ is the total seconds of the output video. 

The resulting transformed coordinates $(x',y',z',t)$ are then processed with positional embedding and passed to the Velocity Network.

\subsection{External Force Network}
Using the feature maps produced by our feature network, we utilize an additional convolutional network to predict external forces for each image. This external force network comprises 7 convolution layers and 4 fully connected layers. The feature dimensions of the convolutional layers are 64, 32, 32, 32, 32, 32, and 32, respectively, while the fully connected layers have feature dimensions of 1024, 256, and 64. The convolution layers include residual connections, with BatchNorm applied to each layer. ReLU activation is used for all layers throughout the network. We set the kernel size to 3 and the stride to 2 for all convolution layers. 

\begin{figure*}[ht]
  \begin{subfigure}{\linewidth}
    \includegraphics[width=0.48\linewidth]{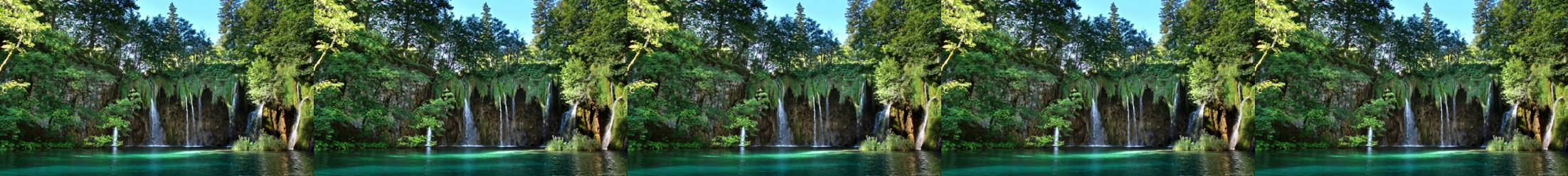}
    \includegraphics[width=0.48\linewidth]{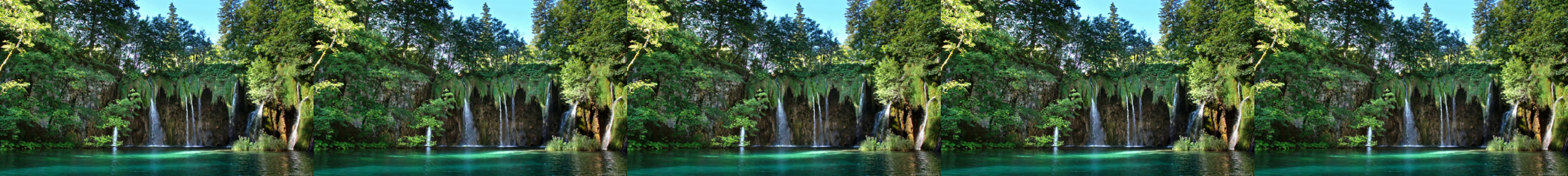}
    \vspace{-3pt}
    \caption{Comparison to ours with viscosity. }
  \end{subfigure}
  \begin{subfigure}{\linewidth}
    \includegraphics[width=0.48\linewidth]{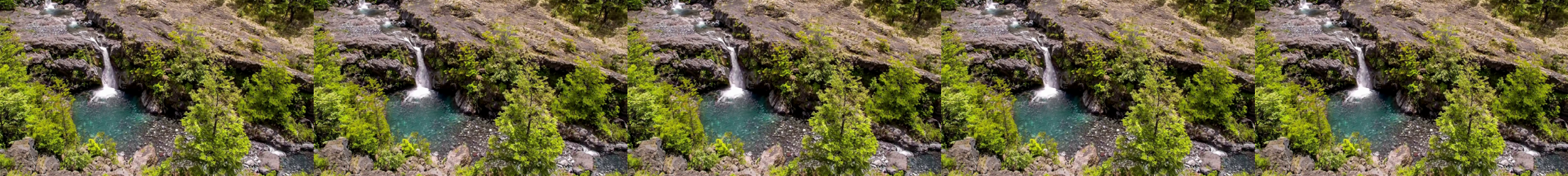}
    \includegraphics[width=0.48\linewidth]{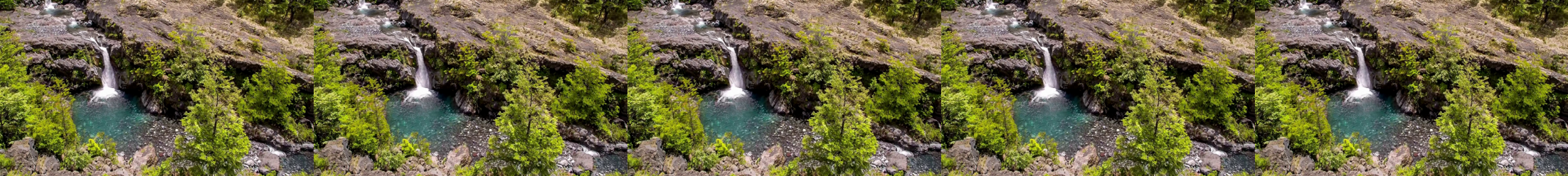}
    \vspace{-3pt}
    \caption{Comparison to ours with pressure fields. }
  \end{subfigure}
   \begin{subfigure}{\linewidth}
    \includegraphics[width=0.48\linewidth]{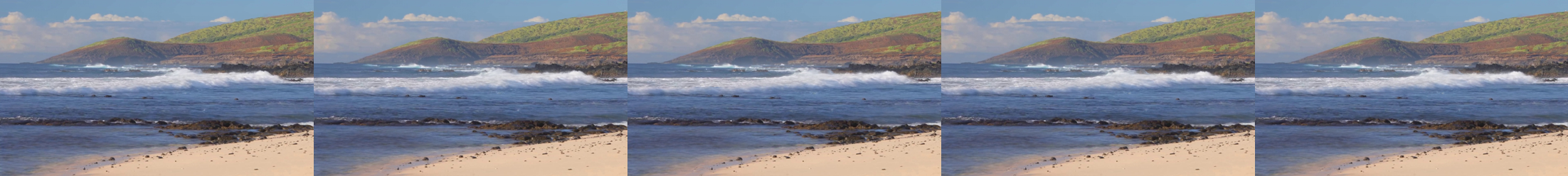}
    \includegraphics[width=0.48\linewidth]{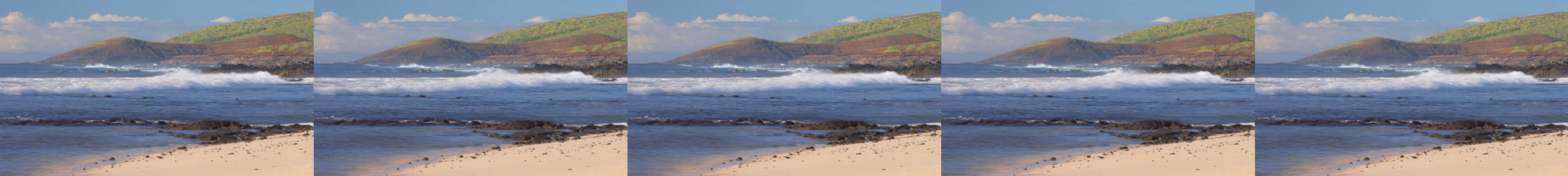}
    \vspace{-3pt}
    \caption{Comparison to ours without external forces. }
  \end{subfigure}
  \vspace{-20pt}
  \caption{Qualitative ablations for Navier-Stokes equation simplification}
  \label{fig:phy_qual}
\end{figure*}

\section{Additional Results}

\subsection{Video Results}
We include video results generated by our methods and baseline methods in the supplementary materials. These results comprise: (a) video outputs from the Holynski et al. ~\cite{holynski} validation set, showcasing both original and novel viewpoints, presented alongside comparisons with baseline methods; (b) video results from edited images, compared with 3D-Cinemagraphy~\cite{3d-cinemagraphy}; and (c) further comparisons with Stable Video Diffusion(SVD)-XT ~\cite{SVD}and CogVideoX-5B~\cite{yang2024cogvideox} on the Holynski \etal~\cite{holynski} validation set.

\subsection{Comparison with diffusion-based methods}
We include several videos generated by SVD-XT~\cite{SVD} and CogVideoX-5B~\cite{yang2024cogvideox} using images from the Holynski \etal~\cite{holynski} validation set. As demonstrated in the video results, diffusion-based methods struggle with complex fluid motion, leading to unreasonable fluid motions and noticeable jittering artifacts in animations.

\subsection{Image Editing}
An image editing example is included in the supplementary material. In this example, we add a stone to the seashore, in which we expect the stone to block and separate the waves. As shown in the video results, our method accurately simulates the intended behavior, with the waves splitting around the stone. In contrast, 3D-Cinemagraphy \cite{3d-cinemagraphy} fails to account for the occlusion, resulting in waves unrealistically passing beneath the stone. Additionally, the results from 3D-Cinemagraphy exhibit other noticeable artifacts, such as holes and dot patterns.

\begin{table}
    \small
    \centering
\resizebox{\linewidth}{!}{
    \begin{tabular}{|c|c|cc|c|}
        \hline
         & Method & SA↑ & PC↑ & Time(s)↓\\
        \hline
        \multirow{3}{*}{Input} & 3D-C & 0.8742 & 0.6936 & 91 \\
         &  Holynski \etal & \textbf{0.8900} & \textbf{0.7171} & 16\\
         & Ours & 0.8739 & 0.7110 & \textbf{13}\\
        \hline
         \multirow{3}{*}{Novel} &  3D-C & 0.8638 & 0.6622 & 90 \\
         &  Make-it-4D & \textbf{0.8702} & 0.6753 & \textbf{24}\\
         & Ours & 0.8618 & \textbf{0.6814} & 29 \\
         \bottomrule    
    \end{tabular}
}
\caption{VideoPhy scores and the computation cost for all methods.}
\label{tab:video_phy}
\end{table}

\subsection{Qualitative ablations for Navier-Stokes equation simplification}
In our ablation study, we present a quantitative analysis supporting our choice of physics guidance. Here, we further provide qualitative ablations in \Cref{fig:phy_qual} to illustrate the visual artifacts produced by alternative designs. Corresponding video results are also available on our project website.

\subsection{VideoPhy Scores}
Recent studies have introduced metrics to directly evaluate how closely generated videos adhere to physical common sense. In \Cref{tab:video_phy}, we report results using the VideoPhy metric~\cite{bansal2024videophy}. As shown in \Cref{tab:video_phy}, the scores across all methods are very similar, suggesting that the current VideoPhy metric may lack the sensitivity to capture fine-grained physical inconsistencies in generated videos.

\subsection{Computation Cost}
Computation cost is also an important consideration for downstream applications. We report the generation time for a 60‑frame 1280×720 video on an H100 GPU in \Cref{tab:video_phy}.

\section{Limitations}
Our method focuses on natural fluids and may be less effective in some scenarios. In particular, the lack of pressure fields limits its ability to capture interactions such as river merging. Static fluid masks also pose challenges for scenes with evolving boundaries. Novel-view videos may contain artifacts due to depth prediction inaccuracies. In the future, we will try to eliminate the Navier-Stokes simplifications and extend our method to compressible fluids by collecting data with pressure, viscosity, and density measurements. We will add this to the paper.

\end{document}